\documentclass[review, 3p]{elsarticle}

\usepackage{amsmath}
\usepackage{multirow}
\usepackage{textcomp}
\usepackage{enumerate}
\usepackage{amssymb}
\usepackage{lscape}
\usepackage{rotating}
\usepackage{tikz}
\usepackage{hyperref}
\usepackage{algorithm}
\usepackage[noend]{algpseudocode}
\usepackage{algorithmicx}
\usepackage{rotating}
\usepackage{booktabs}
\usepackage{tabularx}
\usepackage[acronym, nonumberlist]{glossaries}
\usepackage[skip=2pt]{caption}
\usepackage{subcaption}
\usepackage{tabularx}
\usepackage{afterpage}
\usepackage{makecell}
\usepackage{fancyvrb}
\usepackage{pdflscape} 

\usepackage{lineno,hyperref}
\modulolinenumbers[5]

\makenoidxglossaries

\newacronym{ml}{ML}{Machine Learning}
\newacronym{hp}{HP}{hyperparameter}
\newacronym{dl}{DL}{Deep Learning}
\newacronym{svm}{SVM}{Support Vector Machine}
\newacronym{mtl}{MtL}{Meta-learning}
\newacronym{gs}{GS}{Grid Search}
\newacronym{rs}{RS}{Random Search}
\newacronym{smbo}{SMBO}{Sequential Model-based Optimization}
\newacronym{openml}{OpenML}{Open Machine Learning}
\newacronym{cv}{CV}{Cross-validation}
\newacronym{loocv}{LOO-CV}{Leave-one-out Cross-validation}
\newacronym{nnsmfo}{NN-SMFO}{Nearest Neighbor Sequential Model-Free Optimization}
\newacronym{nmse}{NMSE}{Normalized Mean Squared Error}
\newacronym{knn}{kNN}{k-Nearest Neighbors}
\newacronym{acc}{Acc}{Accuracy}
\newacronym{at}{AT}{Active Testing}
\newacronym{pmcc}{PMCC}{Pearson Product-Moment Correlation Coefficient}
\newacronym{nae}{NAE}{Normalized Absolute Error}
\newacronym{mad}{MAD}{Mean Absolute Deviation}
\newacronym{pso}{PSO}{Particle Swarm Optimization}
\newacronym{ts}{TS}{Tabu Search}
\newacronym{ga}{GA}{Genetic Algorithm}
\newacronym{auc}{AUC}{Area Under the ROC curve}
\newacronym{bac}{BAC}{Balanced per class Accuracy}
\newacronym{cart}{CART}{Classification and Regression Tree}
\newacronym{dt}{DT}{Decision Tree}
\newacronym{dtia}{DTIA}{Decision Tree Induction Algorithm}
\newacronym{nb}{NB}{Na{\"i}ve-Bayes}
\newacronym{rf}{RF}{Random Forest}
\newacronym{lr}{LR}{Logistic Regression}
\newacronym{gp}{GP}{Gaussian Process}
\newacronym{sfs}{SFS}{Sequential Forward Selection}
\newacronym{cd}{CD}{Critical Difference}
\newacronym{eda}{EDA}{Estimation of Distribution Algorithm}
\newacronym{uci}{UCI}{University of California Irvine}
\newacronym{pca}{PCA}{Principal Component Analysis}
\newacronym{fp}{FP}{False Positive}
\newacronym{fn}{FN}{False Negative}
\newacronym{cfs}{CFS}{Correlation-based feature selection}
\newacronym{rbf}{RBF}{Radial Basis Function}
\newacronym{smote}{SMOTE}{Synthetic Minority Over-sampling Technique}
\newacronym{ds}{DS}{Decision Stump}
\newacronym{svd}{SVD}{Singular-value Decomposition}
\newacronym{taf}{TAF}{Transfer Acquisition Function}
\newacronym{irace}{Irace}{Iterated F-race}
\newacronym{roar}{ROAR}{Random Online Adaptive Racing}
\newacronym{pils}{ParamILS}{Iterated Local Search in Parameter Configuration Space}
\newacronym{scrr}{SpCorr}{Spearman Correlation}
\newacronym{mlp}{MLP}{Multilayer Perceptron}
\newacronym{cash}{CASH}{Combined Algorithm Selection and Hyperparameter Optimization}
\newacronym{rl}{RL}{Relative Landmarking}
\newacronym{sm}{SM}{Simple}
\newacronym{st}{ST}{Statistical}
\newacronym{in}{IN}{Information-theoretic}
\newacronym{mb}{MB}{Model-based}
\newacronym{lm}{LM}{Landmarking}
\newacronym{dc}{DC}{Data Complexity}
\newacronym{cn}{CN}{Complex Network}
\newacronym{automl}{AutoML}{Automated Machine Learning}
\newacronym{ela}{ELA}{Exploratory Landscape Analysis}
\newacronym{fosc}{FOSC}{Framework for Optimal Extraction of Clusters}
\newacronym{hdbscan}{HDBSCAN}{Hierarchical Density-Based Spatial Clustering of Applications with Noise}
\newacronym{gbm}{GBM}{Gradient Boosting Machine}
\newacronym{smac}{SMAC}{Sequential Model-based Algorithm Configuration}
\newacronym{scot}{SCoT}{Surrogate Collaborative Tuning}
\newacronym{mklgp}{MKL-GP}{Gaussian Process with Multi Kernel Learning}
\newacronym{rcgp}{RC-GP}{Rank Correlation based Gaussian Process}

\usetikzlibrary{decorations.pathmorphing}

\definecolor{cadmiumgreen}{rgb}{0.0, 0.42, 0.24}
\definecolor{cadmiumred}{rgb}{0.89, 0.0, 0.13}

\newtheorem{definition}{Definition}[section]

\begin{document}

\begin{frontmatter}

\title{Rethinking Default Values: a Low Cost and\\Efficient Strategy to Define Hyperparameters}

\author[utfpr]{Rafael G. Mantovani\corref{mycorrespondingauthor}}
\ead{rafaelmantovani@utfpr.edu.br}
\cortext[mycorrespondingauthor]{Rafael Gomes Mantovani \\ Federal Technology University - Paran\'a, Campus of Apucarana \\
R. Marc{\'i}lio Dias, 635 - Jardim Para{\'i}so, Apucarana - PR, Brazil, Postal Code 86812-460}

\author[unesp]{ Andr\'{e} L. D. Rossi}
\ead{alrossi@itapeva.unesp.br}

\author[usp]{Edesio Alcoba\c{c}a}
\ead{edesio@usp.br}

\author[ufop]{Jadson Castro Gertrudes}
\ead{jadson.castro@ufop.edu.br}

\author[uel]{\\Sylvio {Barbon Junior}}
\ead{barbon@uel.br}

\author[usp]{Andr{\'e} C. P. L. F. {de Carvalho}}
\ead{andre@icmc.usp.br}

\address[utfpr]{Federal Technology University - Paran\'a (UTFPR), Campus of Apucarana - PR, Brazil}
\address[usp]{Institute of Mathematics and Computer Sciences (ICMC), University of S\~ao Paulo (USP), S\~ao Carlos - SP, Brazil}
\address[unesp]{S\~{a}o Paulo State University (UNESP), Campus of Itapeva - SP, Brazil}
\address[ufop]{Department of Computing, Federal University of Ouro Preto (UFOP), Ouro Preto - MG, Brazil}
\address[uel]{State University of Londrina (UEL), Londrina - PR, Brazil}


\begin{abstract}

Machine Learning (ML) algorithms have been increasingly applied to problems from several different areas. Despite their growing popularity, their predictive performance is usually affected by the values assigned to their hyperparameters (HPs). As consequence, researchers and practitioners face the challenge of how to set these values.
Many users have limited knowledge about ML algorithms and the effect of their HP values and, therefore, do not take advantage of suitable settings. They usually define the HP values by trial and error, which is very subjective, not guaranteed to find good values and dependent on the user experience. 
Tuning techniques search for HP values able to maximize the predictive performance of induced models for a given dataset, but have the drawback of a high computational cost.
Thus, practitioners use default values suggested by the algorithm developer or by tools implementing the algorithm.
Although default values usually result in models with acceptable predictive performance, different implementations of the same algorithm can suggest distinct default values. 
To maintain a balance between tuning and using default values, we propose a strategy to generate new optimized default values. 
Our approach is grounded on a small set of optimized values able to obtain predictive performance values better than default settings provided by popular tools.
After performing a large experiment and a careful analysis of the results, we concluded that our approach delivers better default values.
Besides, it leads to competitive solutions when compared to tuned values, making it easier to use and having a lower cost. We also extracted simple rules to guide practitioners in deciding whether to use our new methodology or a HP tuning approach.

\end{abstract}


\end{frontmatter}




\section{Introduction}
\label{sec:intro}

The last decades have seen an explosion of \acrfull{ml} studies and applications, promoting and democratizing its usage by people from diverse scientific and technological backgrounds. Progresses in this area have enabled a large uptake of solutions by the industry and research communities.
Building an \acrshort{ml} solution requires different decisions, and one of them is to choose a suitable algorithm to solve the problem at hand. Different problems present different characteristics, and as a consequence, require different \acrshort{ml} algorithms.

Most of these algorithms have \acrfullpl{hp} whose values directly influence their biases, and consequently, the predictive performance of the induced models. 
Although the number of \acrshort{ml} tools available and their popularity has increased, users still struggle to define the best \acrshort{hp} settings. 
This is not a straightforward task and may mislead practitioners to choose one algorithm over another. Usually, users adjust the \acrshort{hp} values by trial and error, i.e., they empirically evaluate different settings and select what appear to be the best of them.

Ideally, the \acrshort{hp} values should be defined for each problem~\citep{Bergstra:2013b,Reif:2014,Padierna:2017}, trying to find the (near) best settings through an optimization process. As a consequence, several tuning techniques have been used for this purpose. The most simple, and often used, 
are \acrfull{gs} and \acrfull{rs}~\citep{Bergstra:2012}. The former is more suitable for low dimensional problems, i.e., when there are few \acrshortpl{hp} to set. For more complex scenarios, \acrshort{gs} is unable to explore finer promising regions due to the large hyperspace.
The latter is able to explore any possible solution of the hyperspace, but also does not perform an informed search, which may lead to a high computational cost.

Meta-heuristics have also been used for \acrshort{hp} tuning, having the advantage of performing informed searches. Population-based methods, such as~\acrfullpl{ga}~\citep{Ansotegui:2015}, \acrfull{pso}~\citep{Guo:2008} and \acrfullpl{eda}~\citep{Padierna:2017}, have been largely explored in the literature due to their faster convergence. \acrfull{smbo}~\citep{Snoek:2012} is a more recent technique that has drawn a large deal of attention, mainly due to its probabilistic nature. It replaces the target function (\acrshort{ml} algorithm) by a surrogate model~\citep{Bergstra:2011}, which is faster to compute. However, \acrshort{smbo} itself has many \acrshortpl{hp} and does not eliminate the shortcoming of having to iteratively evaluate the function to be optimized.  
All these techniques are valuable alternatives to \acrshort{gs} and \acrshort{rs}, but they might have
a high computational cost, since a large number of candidate solutions usually needs to be evaluated.

A computationally cheaper alternative is to use the default \acrshort{hp} setting suggested by most of the \acrshort{ml} tools. 
These settings may be fixed a priori, regardless of the problem, or defined according to some simple characteristics of the data under analysis.
For instance, the default values of the number of variables selected for each split of the \acrfull{rf} algorithm~\citep{Breiman:2001} and the width of the Gaussian kernel of \acrfullpl{svm}~\citep{Chang:2011} are usually defined based on the number of  
predictive features.

If on the one hand, default values reduce the subjectivity in the experiments, on the other they may not be suitable for every problem~\citep{Braga:2013}, i.e., there is no guarantee that they can result in models with high predictive performance for all cases. 
Besides, many \acrshort{ml} tools follow the same recommendations to propose default settings. Thus, users are not able to find different options using these tools.

Therefore, an alternative that has not received considerable attention in the literature is to consider a pool of default \acrshort{hp} settings, which has a much lower cost than \acrshort{hp} tuning and is less subjective than trial and error.
In practice, instead of trying only one default setting for the HPs, a small promising varied set of \acrshort{hp} settings could be assessed, which are likely to induce models that have better predictive performance than that traditional defaults, using very few evaluations. 
These settings could be acquired from previous experiments of \acrshort{hp} optimization with a varied of datasets.


Hence, in this study, we propose and evaluate a strategy to generate a new set of default \acrshort{hp} settings for \acrshort{ml} algorithms by tuning these values across several datasets. 
We hypothesize that a pool of settings may improve the performance of a model when compared to using only a default setting provided by the \acrshort{ml} tools, with a computation cost much lower than optimization methods. 
The experiments described in this paper evaluated \acrshortpl{svm} due to their well-known hyperparameter sensitivity. However, the whole process can be easily adapted and applied to other~\acrshort{ml} algorithms. 
The process will benefit, especially, algorithms that are sensitive to the choice of their \acrshort{hp} values.


We can summarize the main contributions of this work as:
\begin{itemize}
   
    \item Framing the simple optimization strategy to generate new default \acrshort{hp} settings;
 
    \item Tracing the benefits of multiple default \acrshort{hp} settings by evaluating them across different data domains, and;
 
    \item Performing an in-depth analysis for classification problems, leading to holding discussions and prospecting meta-analyses.
    
\end{itemize}


This paper is structured as follows: Section~\ref{sec:hpt} contextualizes the \acrshort{hp} tuning problem and presents the related work. 
Section~\ref{sec:meth} describes our experimental methodology, detailing how we evaluated the proposed method. The experimental results are discussed in Section~\ref{sec:results}.
Section~\ref{sec:validity} presents possible threats to the validity of the experiments.
The last section draws the conclusions and future work directions.


\section{Hyperparameter Tuning}
\label{sec:hpt}

In a predictive task, \acrfull{ml} algorithms are trained on labeled data to induce a predictive model able to identify the label of new, previously unseen instances.
These algorithms have free ``\textit{\acrfullpl{hp}}'' whose values directly affect the predictive performance of the models induced by them. 
Finding a suitable setting of \acrshort{hp} values requires specific knowledge, intuition, and often trial and error experiments. 
Several \acrshort{hp} tuning techniques, ranging from simple to complex, can be found in the literature.

From a theoretical point of view, selecting the ideal \acrshort{hp} values requires an exhaustive search over all possible subsets of \acrshort{hp} values. 
The number and type of \acrshortpl{hp} can make this task unfeasible.
Therefore, the \acrshort{ml} community usually accepts computing techniques to search for \acrshort{hp} values in a reduced \acrshort{hp} space, instead of the complete space~\citep{Bergstra:2012}. 

Using of computing techniques for \acrshort{hp} tuning has several benefits, such as~\citep{Bardenet:2013}:
\begin{itemize}
    
    \item Freeing the users from the task of manually selecting \acrshort{hp} values, thus they can concentrate efforts on other aspects relevant to the use of \acrshort{ml} algorithms; and
    
    \item Improving the predictive performance of the induced models.
\end{itemize}

Next, we briefly describe the main aspects of \acrshort{hp} tuning, its definition, the main techniques explored in the literature, and related works that are similar to the proposed strategy.


\subsection{Formal Definition}

The \acrshort{hp} tuning process is usually treated as a black-box optimization problem whose objective function is associated with the predictive performance of the model induced by an \acrshort{ml} algorithm. Formally, it can be defined as:

\begin{definition}
Let ${H}={H}_1\times{H}_2\times\dots\times{H}_k$ be the HP space for an algorithm $a\in{A}$, where ${A}$ is a set of ML algorithms. Each ${H}_i$, $i\in\{1, \dots, k\}$, represents a set of possible values for the $i^{th}$ \acrshort{hp} of $a$, and can be defined to maximize
the generalization ability of the induced model.
\end{definition}

\begin{definition}
Let ${D}$ be a set of datasets where $\mathbf{d}\in{D}$ is a dataset from ${D}$.
The function $f:{A}\times{D}\times{H}\rightarrow\mathbb{R}$ measures the predictive performance of the model induced by the algorithm $\mathbf{a}\in{A}$ on the dataset $\mathbf{d}\in{D}$ given a \acrshort{hp} setting $\mathbf{h}=(h_1,h_2,\dots, h_k)\in{H}$. Without loss of generality, higher values of $f(\mathbf{a},\mathbf{d},\mathbf{h})$ mean higher predictive performance.
\end{definition}

\begin{definition}
Given $a\in{A}$, ${H}$ and $\mathbf{d}\in{D}$, together with the previous definitions, the goal of a HP tuning task is to find $\mathbf{h}^\star=(h_1^\star,h_2^\star,\dots, h_k^\star)$ such that
\begin{equation}
\label{eq:setup}
  \mathbf{h}^\star = \underset{\mathbf{h}\in{H}}{arg\:max}\:f(\mathbf{a},\mathbf{d},\mathbf{h})
\end{equation}
\end{definition}

\noindent
The optimization of the \acrshort{hp} values can be based on any performance measure $f$, which can even be defined by multi-objective criteria. Further aspects can make the tuning more complex, such as:
\begin{itemize}
    \item \acrshort{hp} settings that lead to a model with high predictive performance for a given dataset may not lead to high predictive performance for other datasets;
    \item \acrshort{hp} values often depend on each other\footnote{This is the case of~\acrfullpl{svm}~\citep{BenHur:2010}.}. Hence, independent tuning of \acrshortpl{hp} may not lead to a good set of \acrshort{hp} values;
    \item The exhaustive evaluation of several \acrshort{hp} settings can be very time-consuming.
\end{itemize}


\subsection{Tuning techniques}

Over the last decades, different \acrshort{hp} tuning techniques have been successfully applied to
\acrshort{ml} algorithms~\citep{Birattari:2010,Bergstra:2011,Hauschild:2011,Snoek:2012,Bardenet:2013,Li:2018}. 
Some of these techniques iteratively build a population $\mathcal{P}\subset\mathcal{H}$ of \acrshort{hp} settings, when $f(a,\mathbf{d},\mathbf{h})$ is computed for each $\mathbf{h}\in\mathcal{P}$. By doing so, they can simultaneously explore different regions of a search space. 
There are various population-based \acrshort{hp} tuning strategies, which differ in how they update $\mathcal{P}$ at each iteration. Some of them are briefly described next.


\subsubsection{Random Search} 
\label{sub:rs}

\acrfull{rs}~\citep{Andradottir:2015} is a simple technique that performs random trials in a search space. Its use can reduce the computational cost when there is a large number of possible settings being investigated. Usually, \acrshort{rs} performs its search iteratively in a predefined number of iterations. 
$P(i)$ is extended (updated) by a randomly generated \acrshort{hp} setting $\mathbf{h}\in{H}$ in each ($i$th) iteration of the \acrshort{hp} tuning process. \acrshort{rs} has been successfully used for \acrshort{hp} tuning of \acrfull{dl} algorithms~\citep{Bardenet:2013, Bergstra:2012}. 


\subsubsection{Bayesian Optimization}

\begin{sloppypar}
\acrfull{smbo}~\citep{Brochu:2010,Snoek:2012} is a sequential technique grounded on the statistics literature related to experimental design for global continuous function optimization \citep{Hutter:2011}. Many different models can be used inside of \acrshort{smbo}, e.g., Gaussian processes and tree-based approaches. The first provides good predictions in low-dimensional numerical input spaces, allowing the computation of the posterior Gaussian process model. The second, on the other hand, is suitable to address high-dimensional and partially categorical input spaces.

In our particular implementation, \acrshort{smbo} starts with a small initial population $P(0)\neq\emptyset$ which, at each new iteration $i>0$, is extended by a new \acrshort{hp} setting $\mathbf{h}'$, such that the expected value of $f(a,\mathbf{d},\mathbf{h}')$ is maximal. Taking advantage of an induced meta-model $\hat{f}$ approximating $f$ on the current population $P(i-1)$, optimized versions of $\mathbf{h}'$ are intensified until total time budget for configuration is exhausted. In experiments reported in the literature~\citep{Bergstra:2011, Snoek:2012, Bergstra:2013b}, \acrshort{smbo} performed better than \acrshort{gs} and \acrshort{rs} and either matched or outperformed state-of-the-art techniques in several \acrshort{hp} optimization tasks. 
\end{sloppypar}


\subsubsection{Meta-heuristics}

Bio-inspired meta-heuristics are optimization techniques based on biological processes. They also have \acrshortpl{hp} to be tuned~\citep{Reif:2014}. For example, \acrfull{ga}, one of the most widely used, requires an initial population $\mathcal{P}_0=\{\mathbf{h}_1,\mathbf{h}_2,\dots,\mathbf{h}_{n_0}\}$, which can be defined in different ways, and \acrshort{hp} values for operators based on natural selection and evolution, such as crossover and mutation.

Another bio-inspired technique, \acrfull{pso} is based on the swarming and flocking behavior of particles~\citep{Simon:2013}. 
Each particle $\mathbf{h}\in\mathcal{P}_0$ is associated with a position $\mathbf{h}=(h_1,\dots,h_k)\in\mathcal{H}$ in the search space $\mathcal{H}$, a velocity $\mathbf{v}_h\in\mathbb{R}^k$ and the best position found so far $\mathbf{b}_h\in\mathcal{H}$.
During its iterations, the movement of each particle is changed according to its current best-found position and the current best-found position $\mathbf{w}\in\mathcal{H}$ of the entire swarm (recorded through the
optimization process).

\begin{sloppypar}
Another popular technique, \acrfull{eda}~\citep{Hauschild:2011},
combines aspects of \acrshort{ga} and \acrshort{smbo} to guide the search by iteratively updating an explicit probabilistic model of promising candidate solutions. For such, the implicit crossover and mutation operators used in \acrshort{ga} are replaced by an explicit probabilistic model $M$.
\end{sloppypar}


\subsubsection{Iterated F-Race} 
\label{sub:irace}

The \acrfull{irace}~\citep{Birattari:2010} technique was designed to use `\textit{racing}' concepts for algorithm configuration and optimization problems~\citep{Lang:2013, Miranda:2014}. One race starts with an initial population $\mathcal{P}_{0}$, and iteratively selects the most promising candidates considering the distribution of \acrshort{hp}  values, and 
statistical tests. Configurations (settings) that are statistically worse than at least one of the other configuration candidates are discarded from the racing. Based on the surviving candidates, the distributions are updated. This process is repeated until a stopping criterion is reached.


\subsection{Related Works}
\label{subsec:related}

\begin{figure}[h!]
    \centering
    \includegraphics
    [scale=0.65]{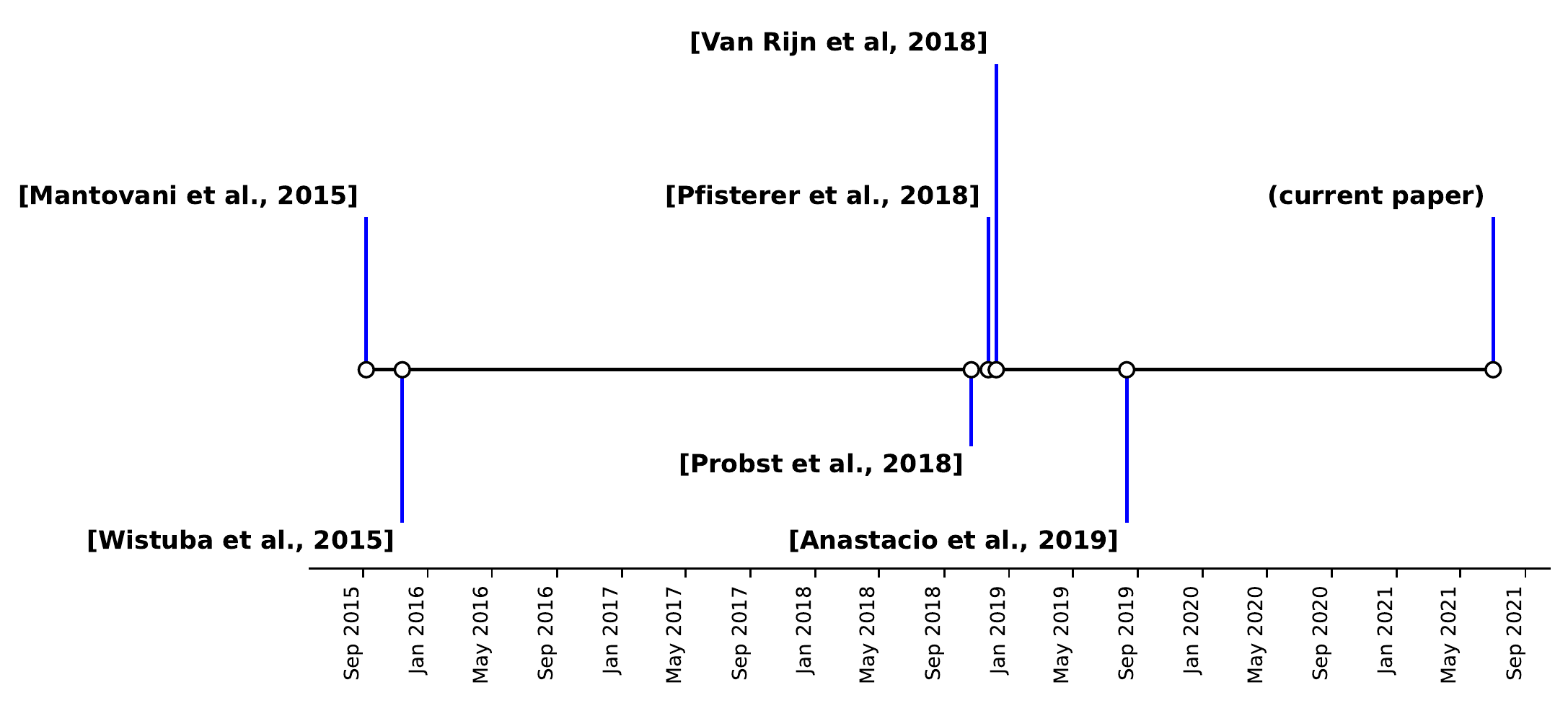}
    \caption{Timeline of related studies which perform automatic design of defaults hyperparameter values.}
    \label{fig:timeline}
\end{figure}

In our literature review, we found a small number of studies investigating the automatic design of default \acrshort{hp} values. Figure~\ref{fig:timeline} 
summarizes the related studies according to their publication date.
The research question itself is very recent, motivated by the high number of public experimental results made available by the \acrshort{ml} research community~\footnote{A high number of experimental results can be obtained from the OpenML website: \url{https://www.openml.org/search?type=run}.}. The first two investigations of the design of \acrshort{hp} settings were published in 2015~\citep{Mantovani:2015-def, wistuba2015}, with the remaining developments concentrated in the years 2018-2019~\citep{probst2018, Pfisterer:2018, Rijn:2018, Anastacio:2019}. 
We detail the main aspects of these works into tables: Table~\ref{tab:related_1} presents the \acrshort{ml} algorithms and datasets investigated by each related work; while Table~\ref{tab:related_2} shows the methodology adopted to perform the \acrshort{hp} optimization. 
 optimization. 


\begin{table*}[hp!]
\scriptsize
\centering
\caption{Main characteristics of the ``\textit{learning}'' task performed by related studies. Columns show for each related study: its reference, year of publication, \acrshort{ml} algorithms studied, and the number/source of the datasets.} 
\label{tab:related_1}
\setlength{\tabcolsep}{7pt}
\begin{tabular}{llrccc}
    \toprule
    \multirow{2}{*}{\textbf{Reference}} &  \multirow{2}{*}{\textbf{ML Algorithms}} & \multirow{2}{*}{\textbf{\# Datasets}} & \multicolumn{3}{c}{\textbf{Data Source}} \\
    & & & \textbf{\acrshort{openml}} & \textbf{\acrshort{uci}} & \textbf{AClib} \\
    \midrule
    
    \rule{0pt}{1ex}
    \cite{Mantovani:2015-def} & \acrshort{svm} & 145\:\: &  $\bullet$ & $\bullet$ \\
    
    \rule{0pt}{5ex}
    \cite{wistuba2015} &  \acrshort{svm}, AdaBoost & 25\:\: & & $\bullet$ \\
    
    \rule{0pt}{5ex}
    
    \multirow{2}{*}{\cite{probst2018}} & \acrshort{svm}, \acrshort{knn}, \acrshort{cart}, & \multirow{2}{*}{38\textbf{*}} & \multirow{2}{*}{$\bullet$} \\
    & \acrshort{gbm}, \acrshort{rf}, Elastic-net \\
    
    \rule{0pt}{5ex}
    \multirow{2}{*}{\cite{Pfisterer:2018}} & \acrshort{svm}, AdaBoost, \acrshort{cart}, & \multirow{2}{*}{38\textbf{*}} & \multirow{2}{*}{$\bullet$} & \\
    & \acrshort{gbm}, \acrshort{rf}, Elastic-net \\
    
    \rule{0pt}{5ex}
    \cite{Rijn:2018} & \acrshort{svm} & 98\:\: & $\bullet$ \\
    
    \rule{0pt}{5ex}
    \cite{Anastacio:2019} &  Auto-WEKA & 20\:\: & & & $\bullet$\\   
    
    \bottomrule
    \end{tabular}
    
    
    \begin{flushleft}
    *Only binary classification problems.\\
    \end{flushleft}
    \vspace{-1cm}
    
    \vspace{1.5cm}
    \centering
    \caption{Main characteristics of the ``\textit{tuning}' task performed by related studies. Columns show for each related study: its reference, year of publication, tuning techniques explored, performance measures used, the evaluation methodology and baselines used in experimental comparisons.} 
    \label{tab:related_2}
    \setlength{\tabcolsep}{5pt}
    \begin{tabular}{lllll}
        \toprule
        \multirow{2}{*}{\textbf{Reference}} &
        \multicolumn{1}{l}{\textbf{Tuning}} & \textbf{Performance} & \textbf{Evaluation} & \multirow{2}{*}{\textbf{Baselines}} \\
        & \multicolumn{1}{l}{\textbf{Techniques}} & \textbf{Measures} & \textbf{Procedures} \\
        \midrule
        
        \rule{0pt}{1ex}
        \multirow{3}{*}{\cite{Mantovani:2015-def}}  & 
        \multirow{3}{*}{\acrshort{pso}} & \multirow{3}{*}{\acrshort{bac}} & Nested-\acrshort{cv} & WEKA defaults \\
         & & & Outer: 10-\acrshort{cv} & LibSVM defaults \\
         & & & Inner: Holdout &  \\

        \rule{0pt}{5ex}
        \multirow{3}{*}{\cite{wistuba2015}} &
        \multirow{3}{*}{NN-SMFO} & Ranking & Nested-\acrshort{cv} & \acrshort{scot}, \acrshort{rs}, \\
        & & CANE & Outer: 10-\acrshort{cv} (outer) & \acrshort{smac}++\\
        & & & Inner: Holdout & \acrshort{mklgp}, \acrshort{rcgp} \\
    
        \rule{0pt}{5ex}
        \multirow{2}{*}{\cite{probst2018}} & 
        \multirow{2}{*}{\acrshort{smbo}} & Accuracy, \acrshort{auc} & Single-\acrshort{cv} & \multirow{2}{*}{None} \\
        & & $R^2$, Kendall's tau & 10-\acrshort{cv} \\
       
        \rule{0pt}{5ex}
        \multirow{3}{*}{\cite{Pfisterer:2018}} &
        \acrshort{rs} & Accuracy & Nested-CV  & \multirow{3}{*}{None}\\
        & \acrshort{smbo} & \acrshort{auc} & Outer: 10-\acrshort{cv} \\
        & & & Inner: Holdout\\
        
         \rule{0pt}{5ex}
        \multirow{2}{*}{\cite{Rijn:2018}} &\multirow{2}{*}{\acrshort{gs}} & \multirow{2}{*}{\acrshort{auc}} & Single-\acrshort{cv} & \multirow{2}{*}{None} \\
        & & & 10-\acrshort{cv}  \\
        
        \rule{0pt}{5ex}
        \multirow{3}{*}{\cite{Anastacio:2019}} & \acrshort{smac} & \multirow{3}{*}{Error rate} & \multirow{3}{*}{single-CV} & \multirow{3}{*}{Auto-WEKA defaults}\\
        & GGA++ \\
        & \acrshort{irace} \\
    
    \bottomrule
    \end{tabular}
\end{table*}


\subsubsection{Our very first try on shared default settings}

In~\cite{Mantovani:2015-def}, the authors used the \acrfull{pso} algorithm to perform~\acrshort{hp} tuning of \acrshortpl{svm},  and find new ``default'' hyperparameter settings for them. The optimization task was performed simultaneously with $21$ random datasets. The \acrshort{hp} settings returned by the \acrshort{hp} tuning task were considered as new ``optimized default'' \acrshort{hp} settings. These values were compared to the default settings recommended in the \texttt{Weka} tool~\cite{frank2016} and in the \texttt{LibSVM} library~\cite{Chang:2011}. 
The experiments showed promising results whereby the new optimized settings induced models better than baselines for most of the investigated datasets. 

\subsubsection{Sequential Model-Free Hyperparameter Tuning}

In~\cite{wistuba2015}, the authors proposed a method to select the best \acrshort{hp} setting from a finite set of possibilities, which can be seen as a set of default \acrshort{hp} settings. 
They used a \acrfull{nnsmfo} method to assess the average performance of an \acrshort{hp} setting only on the $k$ datasets that were similar to the new test dataset.
Two datasets are considered similar if they behave similarly, e.g., they have similar predictive performance rankings with respect to the \acrshort{hp} settings. 
The authors claim that few evaluations are enough to approximate the true rank and that performance ranking is more descriptive than distance functions based on meta-features, i.e., features describing properties of a dataset, and construct the feature space for meta-learning. 
Experiments were performed with AdaBoost and~\acrshortpl{svm} using a set of 108 and 288~\acrshort{hp} settings, respectively, generated from a grid of \acrshort{hp} values. Besides converging faster than the other strategies, \acrshort{nnsmfo} also presented the smallest ranking and average normalized error. 
However, these gains were not validated by a statistical significance test. In addition, the coarse~\acrfull{gs} used in the experiments was probably missing promising~\acrshort{hp} search space regions.


\subsubsection{Tunability: Importance of Hyperparameters of Machine Learning Algorithms}

\cite{probst2018} also defined default \acrshort{hp} values empirically based on experiments with $38$ binary classification datasets. In their experiments, the best default \acrshort{hp} setting is the configuration that minimizes on average a loss function considering many datasets, i.e., \acrshort{hp} default settings are supposed to be suitable across different datasets. 
In the study, a set of \acrshort{hp} settings is not directly evaluated due to the high computational cost. Instead, a surrogate regression model based on a meta-dataset is used. Thus, the surrogate model learns to map an \acrshort{hp} setting to the estimated performance of a \acrshort{ml} algorithm with respect to a dataset.
The authors performed experiments with six \acrshort{ml} algorithms, including \acrshort{dt} induction, \acrshortpl{svm} and gradient boosting. 
At the end, they analyzed the impact of tuning the algorithm and its \acrshortpl{hp}, namely tunability. 

\subsubsection{Learning Multiple Defaults for Machine Learning Algorithms}

In~\cite{Pfisterer:2018}, the authors wanted to find default values that generalize well for many datasets instead of only for specific datasets. In their experiments, they took advantage of a large database of prior empirical evaluations available on \acrshort{openml}~\citep{Vanschoren:2014} to explore their hypothesis. 
Due to the high computational cost to estimate the expected risk of an induced algorithm using \acrshort{cv}, surrogate models were used to predict the performance of the \acrshort{hp} settings. Thus, any \acrshort{hp} setting is evaluated faster using a surrogate model trained for each dataset. 
Finally, a greedy optimization technique searches through a list of defaults based on the predictions of the surrogate models. They assume that this list has at least one setting that is suitable for a given dataset. 
Experiments were carried out using 6 learning algorithms on a nested leave-one-out \acrshort{cv} resampling method for up to 100 binary balanced datasets. According to the experimental results, a set of at most 32 new default settings outperformed two baseline strategies, \acrshort{rs} and \acrshort{smbo}, for a budget size with 32 and 64 iterations, respectively. Therefore, new default settings are especially valuable when processing time is scarce to perform \acrshort{hp} tuning.

Although the new default values are interesting from the practical point of view, this study was not concerned with the default values found and the characteristics of the datasets. Our current study overcomes this necessity to better understand the problem itself, which may be useful to the proposal of alternative default \acrshort{hp} settings. \acrshort{hp} tuning is usually performed by most of the methods starting from scratch for each dataset. If \acrshort{hp} values are somehow dependent on dataset properties, this relation could be learned for a warm start of optimization methods \citep{Reif:2012,Gomes:2012}, for the prediction of \acrshort{hp} values~\citep{Eggensperger:2018} or for the proposal of symbolic default \acrshort{hp} settings \citep{Rijn:2018}.


\subsubsection{Meta Learning for Default--Symbolic Defaults}

Instead of searching for a good set of default \acrshort{hp} values, \cite{Rijn:2018} used meta-learning to learn sets of symbolic default \acrshort{hp} settings suitable for many datasets. Symbolic default values are functions of the characteristics of the data rather than static values. An example of symbolic default is the relation to the number of features $n$ used by LibSVM to define the width ($gamma$) of the Gaussian kernel ($\gamma = 1/N$) \acrshort{hp}. Experiments were performed considering five different functions (transformations) over 80 meta-features for the $\gamma$ and $C$ of a SVM with Gaussian kernel.  
The experimental results showed that this technique is competitive to \acrfull{gs} using a surrogate model to predict the performance on a specific dataset. 
However, in this study, the authors only analyzed a symbolic default at each time when other \acrshortpl{hp} received static values, i.e., the authors did not take into account how multiple \acrshortpl{hp} could interact. 

\subsubsection{Importance of Default values for a warm-start HP tuning} 

Usually, algorithm configurators initialize their search for the best settings based on random values. An alternative for a warm-start is to take advantage of default settings. According to~\cite{Anastacio:2019}, default settings contain valuable information that can be exploited for \acrshort{hp} tuning. Guided by this hypothesis, they investigated the benefit of using default settings for different automatic configurators, namely
~\acrshort{smac}~\citep{Hutter:2011}, GGA++~\citep{Ansotegui:2015}, and \acrshort{irace}~\citep{Birattari:2010}. In addition, they proposed two simple methods to reduce the search space based on default values. The empirical analysis was performed considering 20 problems of AClib~\citep{Hutter:2014}, including four datasets to evaluate Auto-WEKA, an automated searching system based on the WEKA learning algorithms and their \acrshort{hp} settings. According to experimental results, default hyperparameter settings can critically influence on the configurators' performance. This positive impact was observed mainly for \acrshort{irace} and other \acrshort{ml} problems. Moreover, the methods to reduce the search space led to smaller error rates for the \acrshort{smac} algorithm. Thereby, the authors claim that default hyperparameter settings provide valuable information for automated algorithm configuration. 


\subsection{Summary of Literature Overview}

As previously mentioned, the literature review carried out for this study found only six studies investigating the generation of default \acrshort{hp} settings for \acrshort{ml}, each addressing a related issue, as discussed next:

\begin{itemize}

    \item Three studies did not exactly perform \acrshort{hp} tuning~\citep{wistuba2015, Pfisterer:2018, probst2018}: they ``\textit{simulate}'' \acrshort{hp} tuning via surrogate models. These surrogate models predict the expected performance for a given \acrshort{hp} setting. The benefits of this approach are to set up the optimization process and reduce the cost associated with evaluating each single setting. On 
    the other hand, they can propagate an erroneous value if the predictions are very different from the true predictive performance values;
 
    \item In \cite{Rijn:2018}, the authors try to induce new symbolic relationships between the datasets' characteristics to set the \acrshort{hp} values of an \acrshort{ml} algorithm. In fact, they do not directly suggest a value, but a heuristic (formula) whose output depends on the dataset used;
    
    \item In \cite{Pfisterer:2018}, the authors propose a pool of default \acrshort{hp} settings according to empirical data available on \acrshort{openml}. There is also no tuning, just ranking and evaluation of prior \acrshort{hp} settings. This technique might work in specific conditions, but it is not clear how these default values work for new datasets;
    
    \item In \cite{Anastacio:2019}, the authors investigate the warm start of algorithm configuration tools, but the evaluation is focused on algorithm configuration problems (AClib).
     
\end{itemize}



\section{Experiments}
\label{sec:meth}

In this section we present an overview of the experimental strategy adopted to generate a pool of \acrshort{hp} settings (Sec.~\ref{sec:strategy}), the datasets (Sec.~\ref{sec:datasets}),  and the experiments carried out to evaluate this strategy (Sec.~\ref{sec:opt_process}).
Our hypothesis is that the \acrshort{hp} settings generated by this strategy can lead to models with high predictive performance for datasets of different domains, and, therefore, can be considered as default values by a ML tool.
The main purpose of providing such settings is to save users time, which can evaluate a small number of potential settings, instead of performing \acrshort{hp} tuning for each new dataset under analysis, which is very time consuming.


\subsection{Generating default HP settings}
\label{sec:strategy}

\begin{figure}[htb]
     \centering
     \includegraphics
     [width=0.9\textwidth]{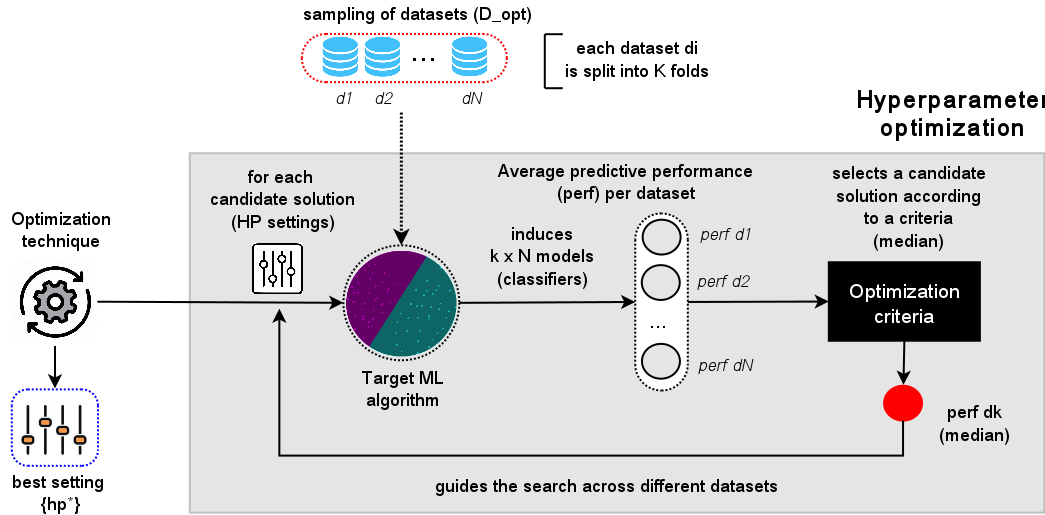}
     \caption{The proposed strategy to generate \acrshort{hp} settings through an optimization process over datasets of different domains.}
     \label{fig:exp-strategy}
\end{figure}

An overview of the experimental strategy followed to generate default \acrshort{hp} settings is presented in Figure~\ref{fig:exp-strategy}. Before explaining this strategy, we need to define some terms and choices:

\begin{itemize}
    \item \textit{a target \acrshort{ml} algorithm}: the algorithm whose hyperparameter values will be optimized; 
    \item \textit{an optimization technique}: a technique that will conduct the optimization process;
    \item \textit{a sample of datasets}: a set of datasets used to evaluate candidate solutions for new default \acrshort{hp} settings; and
    \item \textit{an optimization criterion}: a criterion that defines how candidate solutions will be evaluated and handled during the optimization.
\end{itemize}


Since the search process is guided by the predictive performance of the models induced by a target algorithm, the proposed strategy will generate \acrshort{hp} settings that are restrict to this algorithm. Although this process is computational costly, this is performed only once and the generated settings can be added in the \acrshort{ml} tools as default values for this specific algorithm. Regarding the optimization technique, there is no restriction for this choice and, thus, any general purpose technique can be explored.
However, considering that the goal is to generate a pool of \acrshort{hp} settings, while population-based optimization algorithms are able to generate a pool of candidate solutions at once,other techniques, for the same purpose, need to be run multiple times.
Additionally, this pool should be as diverse as possible, i.e., it is expected that these settings represent different regions of the search space.

With both target and optimization algorithms defined, a sample of  datasets $D$ representing a wide range of different application domains has to be rigorously selected.
In this case, it should be avoided to include, for instance, datasets that were created through data transformation processes from other datasets. 
These datasets are randomly split into two partitions, D$_{opt}$ and  D$_{test}$. The former partition is used during the \acrshort{hp} optimization process to generate a pool of \acrshort{hp} settings, while the latter is used to assess the predictive performance of the \acrshort{ml} technique using these settings. 
This is very similar to a traditional holdout method, but instead of splitting the examples of a dataset, each instance is, itself, a dataset, named $d_i$.

In a traditional \acrshort{hp} tuning design, the optimization technique is usually guided assessing the model's predictive performance for different \acrshort{hp} settings considering a specific dataset at hand.
On the other hand, in the proposed strategy, for each candidate solution $h$ generated by the optimization technique, the target learning algorithm induces a model (classifier) for each dataset $d_i \in D_{opt}$ with $i=\{1,2,3,\ldots,N\}$.
The predictive performance of each model is assessed by the stratified cross-validation method (perf $d_i$) and an appropriate measure. Based on these $N$ performances, the optimization technique uses a criterion, such as mean or median, to determine the fitness value of the candidate solution $h$.

At the end of the optimization process, one or more \acrshort{hp} settings, which best performed, can be adopted
by this strategy. E.g., if a population-based technique is used, either the best individual or the whole population could be considered. In addition, other variability factors could be explored to obtain a pool of settings, such as by choosing different subsets of $D_{opt}$.


\subsection{Datasets} 
\label{sec:datasets}

For the experiments carried out in this study, we used a dataset collection from the \acrfull{openml} repository~\citep{Vanschoren:2014} that were curated by~\cite{Mantovani:2019}.
This collection has $156$ heterogeneous datasets, covering problems from different application domains, presenting different numbers of features, examples, and classes. From this set of datasets, the optimization process could not finish within the defined budget time (100 hours) for $6$ datasets. Therefore, in the present paper, $150$ datasets were used in the experiments.
We considered preprocessing steps to enlarge the number of viable datasets in our experiments without affecting the results, since it occurred before any tuning procedure.

In order to be suitable for \acrshortpl{svm}, all the datasets were preprocessed by:
\begin{itemize}
    \item Removing constant and identifier features;
    \item Converting logical (boolean) attributes into numeric values $\in \{0,1\}$;
    \item Imputing missing values by the median in numerical attributes and a new category for categorical ones;
    \item Converting all the categorical features to numerical values using the 1-N encoding;
    \item Normalizing all the features with $\mu = 0$ and $\sigma=1$.
\end{itemize}

After preprocessing the datasets, they were randomly split into two partitions, $D_{opt}$ and $D_{test}$, of equal sizes ($75$ datasets each). The datasets in $D_{opt}$ are used during the \acrshort{hp} optimization process, while the effect of the settings in the predictive performance are assessed using the datasets in $D_{test}$. 
Since these partitions are defined at random, we repeated the sampling process five ($5$) times, to reduce the variance related to the dataset split~\citep{Cawley:2010}. This is similar to the 5x2 CV, but each instance of a partition represents a dataset itself, instead of an example.

We used the \texttt{mlr}~\citep{mlr:2016}\footnote{\url{https://github.com/mlr-org/mlr}} R package to preprocess the datasets. 
More information regarding datasets and the selection criteria can be found in~\cite{Mantovani:2019}. Besides, a detailed list of these datasets can be found in our \acrshort{openml} study page~\footnote{\url{https://www.openml.org/s/52/data}}.


\subsection{Optimization process}
\label{sec:opt_process}


Although our strategy is suitable to generate a pool of new (default) \acrshort{hp} settings for any \acrshort{ml} algorithm, we decided to evaluate this strategy using \acrshort{svm} as the target algorithm due to its highly sensitive to \acrshort{hp} tuning~\citep{Bergstra:2012,Mantovani:2019}, as confirmed by our literature review (see Table~\ref{tab:related_1}). Thus, new optimized default settings can be useful to avoid \acrshort{hp} tuning and, consequently, reduce the computational cost of running these processes. 


We selected the~\acrfull{pso} as the optimization technique~\citep{Shi:1999} for these experiments. The literature has benchmarked different optimization techniques for \acrshort{svm} tuning~\citep{Mantovani:2018-thesis}. However, they showed similar results when comparing the performance of their induced models. Among the tuning techniques reported, the~\acrshort{pso} converged faster than the others, it did not require prior tuning, and was robust to obtain accurate \acrshort{hp} settings in different types of datasets.


Another decision regarding the experimental strategy is the number of datasets $D_{opt}$ used in the optimization. The smaller the number of datasets, the less the computational time necessary to perform the \acrshort{hp} optimization. 
Besides, there is not a strong assumption that the larger the number of datasets, the better the settings found, since it could prevent the diversity of \acrshort{hp} settings.
Thus, we investigated whether it is possible to obtain suitable settings with few datasets, and how the number of datasets affects the quality of these settings. For such, we evaluated $4$ (four) different sample sizes of $D_{opt}$, $S_k$, with $k=\{11,31,51,71\}$ datasets.
The smallest sample considers just $11$ of the $75$ datasets in $D_{opt}$, while the largest sample, $S_{71}$, includes almost all the $D_{opt}$ datasets. 
It is important to mention that the smallest samples are contained in the largest samples, i.e., $S_{11} \subset S_{31} \subset S_{51} \subset S_{71} \subset D_{opt}$.


The optimization criterion used by the PSO to assess the fitness of a candidate solution $h$ is the median of the predictive performance of $h$ over all datasets in $S_{k}$ measured by the \acrshort{bac} measure. This centrality measure was chosen because it is less sensitive to outliers than the mean measure.


\subsubsection{Hyperparameter space}
\label{sec:hpspace}

The \acrshort{svm} \acrshort{hp} space used in the experiments is presented in Table~\ref{tab:hp-space}. For each \acrshort{hp}, we show its symbol, name, and range or options of values.
We only considered the \acrfull{rbf} kernel in the experiment since: i) it achieves good performance values in general; ii) it may handle nonlinear decision boundaries, and iii) it can approximate the other kernel types~\citep{Hsu:2007}. The \acrshort{hp} ranges shown in this table were first explored in~\cite{Ridd:2014}. 

\begin{table}[h!]
\begin{minipage}{\textwidth}
    \centering
    \caption{\acrshort{svm} hyperparameter space used in experiments. This table shows, for each hyperparameter, its symbol, name and range/options of values when tuned.}
    \label{tab:hp-space}
        \begin{tabular}{clcc}
           \toprule
            \textbf{Symbol} & \textbf{ Hyperparameter } & \textbf{ Range/Options } \\ 
            \midrule
            \rule{0pt}{1ex} 
            k & kernel &\{RBF\} \\ 
            \rule{0pt}{3ex} 
            C & cost & $[2^{-15}, 2^{15}]$ \\
            \rule{0pt}{3ex} 
            $\gamma$ & width of the kernel & $[2^{-15}, 2^{15}]$ \\ 
            \bottomrule
        \end{tabular}
\end{minipage}
\end{table}


\subsubsection{Experimental setup}
\label{sub2:exp_setup}

We present the complete experimental setup in Table~\ref{tab:hp_setup}. 
Since \acrshort{pso} is a stochastic method, we executed it $10$ times with different seeds for each sample $S_k$, with a population of $10$ particles and a budget of $300$ iterations or $100$ hours. The number of evaluations was defined by prior experiments with \acrshort{hp} tuning evaluation~\citep{mantovani:2015-rs}. The initial population was warm-started with the defaults from WEKA ($Cost = 1$, $\gamma = 0.01$). 
To assess the models' performance in the fitness function, we used a single stratified \acrfull{cv} resampling method with $10$ folds. Only the particle (solution) with the best fitness is included in the pool of default \acrshort{hp} settings.

Therefore, in our experimental setup, the proposed strategy is generating a pool of $50$ \acrshort{hp} settings for each sample size $S_k$. This number is due to the 10 repetitions of PSO times 5 replications of the sampling method ($5 \times 2$  \acrshort{cv}, Sec.~\ref{sec:datasets}). Only the best particle of each \acrshort{pso} repetition is included in the pool. \\ 

To evaluate these $50$ settings, every \acrshort{hp} setting in this pool is assessed for every dataset in $D_{test}$, selecting the best setting per dataset, i.e., the \acrshort{hp} setting that induced the model with the highest predictive performance. The results of this evaluation are  hereafter referred to as ``default.opt'' (default optimized \acrshort{hp}), or ``def.opt'' for short. We compared \texttt{default.opt} with two baselines: 
\begin{enumerate}
    
    \item Defaults from \acrshort{ml} tools (lower bound): default \acrshort{hp} values from mlr (LibSVM/R), Weka (Java) and scikit-learn (Python) software/packages; and
    
    \item Conventional \acrshort{hp} tuning (upper bound): \acrshort{hp} tuning results of an \acrshort{rs} technique performed on each dataset with the same budget (300 evaluations). \acrshort{rs} proved to be competitive for \acrshort{svm} assessment~\citep{Bergstra:2012, mantovani:2015-rs}, inducing models as accurate as those induced by more robust techniques (\acrshort{smbo}, \acrshort{pso}, \acrshort{eda}).
    
\end{enumerate}

The predictive performance of the proposed strategy and the baselines are assessed by the \acrshort{bac} measure for the test datasets $D_{test}$ and the $10$-fold stratified \acrshort{cv} resampling procedure. It is important to note that the \acrshort{hp} settings suggested by
\texttt{default.opt} and \acrshort{rs} were estimated using $D_{opt}$ and assessed in $D_{test}$, which are mutually exclusive (Section \ref{sec:datasets}).


Hence, the tuning setup detailed in Table~\ref{tab:hp_setup}  were executed by parallelized jobs in a cluster facility\footnote{\url{http://www.cemeai.icmc.usp.br/Euler/index.html}} and it took one month to be completed.
The \acrshort{pso} algorithm was implemented in R using the \texttt{pso} package\footnote{\url{https://cran.r-project.org/web/packages/pso/index.html}}. 
The code developed for this study is also hosted at \texttt{GitHub}~\footnote{\url{https://github.com/rgmantovani/OptimDefaults}}. There, one can find the optimization jobs and the automated graphical analyses.

\begin{table*}[th!]
\scriptsize
\centering
\caption{Hyperparameter tuning experimental setup.}
\begin{tabular}
    {@{\extracolsep{\fill}}lllc}
\toprule
\textbf{Element} & \textbf{Feature} & \textbf{Value} & \textbf{R package} \\
\midrule
\rule{0pt}{1ex} 

\multirow{7}{*}{\acrshort{hp} tuning} & Technique & Particle Swarm Optimization & \multirow{7}{*}{\texttt{pso}} \\
    & Stopping criteria & budget size \\
    & Population size & 10\\
    & Maximum number of iterations & 30 \\
    & Budget size & 300 \\
    & fitness criteria & median \\
    & Algorithm implementation & SPSO2007$^{1}$ \\

\rule{0pt}{4ex} 
Target algorithm & \acrshort{ml} algorithm & Support Vector Machines & \texttt{e1071} \\

\rule{0pt}{4ex} 
Sample size  & Number of datasets & \{11, 31, 51, 71\} &  \\

\rule{0pt}{4ex} 
\multirow{2}{*}{Resampling Strategy} & \multirow{2}{*}{Single Loop} & \acrfull{cv} & \multirow{2}{*}{\texttt{mlr}} \\
 & & 10-fold & \\

\rule{0pt}{4ex}
\multirow{3}{*}{Performance measures} & Optimized (fitness) & Balanced per class accuracy & \multirow{3}{*}{\texttt{mlr}} \\
 & \multirow{2}{*}{Evaluation} & \{Balanced per class accuracy, \\
 & & optimization paths\} \\

\rule{0pt}{4ex} 
\multirow{2}{*}{Repetitions} & \multirow{2}{*}{Seeds} & 10 values \\
& & seeds = $\{1, \ldots, 10\}$ \\

\rule{0pt}{4ex} 
\multirow{4}{*}{Baselines} & \multirow{3}{*}{Default settings} & LibSVM$^{2}$ & \texttt{e1071} \\
& & WEKA$^{3}$ & \texttt{RWeka} \\
& & Scikit-learn$^{4}$ & - \\
& Tuning & Random Search & \texttt{mlr} \\
\bottomrule
\end{tabular}
 \begin{flushleft}
 1 - Implementation detailed in~\cite{Clerc:2012}.\\
 2 - https://cran.r-project.org/web/packages/e1071/e1071.pdf \\
 3 - https://weka.sourceforge.io/doc.dev/weka/classifiers/functions/SMO.html\\
 4 - https://scikit-learn.org/stable/modules/generated/sklearn.svm.SVC.html\\
 \end{flushleft}
\label{tab:hp_setup}
\end{table*}


\section{Results and Discussion} 
\label{sec:results}

In this section, we present and discuss the main experimental results obtained using the proposed strategy to generate a pool of default \acrshort{hp} settings. 
An overall analysis, including the comparison of the proposed strategy with the baselines, is introduced in Section~\ref{sec:res_opt_def}, while a detailed assessment of these achievements and the optimized default settings found by our strategy are presented in Sections \ref{sec:improvement-analysis} and ~\ref{sec:analsyis_def_opt}, respectively. 
How these results could help us to better understand whether or not \acrshort{hp} tuning should be performed for a give problem is investigated in Section ~\ref{sec:learning_from_def_opt}.
All these results refer to the pool of \acrshort{hp} settings using the sample size $k=51$, which led to the models with the highest predictive performance.
Finally, we discuss and analyze the sensitivity of our strategy to different sample sizes in Section~\ref{sec:sample_size_sensitivity}.


\subsection{Optimized default settings}
\label{sec:res_opt_def}

The results considering all strategies for setting SVM \acrshort{hp}, including default.opt with sample size $k=51$, are presented in Figure~\ref{fig:violin_performance_sample_51} (see Appendix \ref{sec:all-sample-sizes} for other sample sizes). The violin plots show the distribution of the \acrshort{bac} values (x-axis) obtained by these strategies (y-axis) across all datasets in $D_{test}$, sorted accordingly to their average performance. The vertical red dotted line represents the median value obtained by default.opt strategy and is used to highlight differences for the baselines.
Comparing the strategies, the best results were obtained by the \acrshort{rs} technique (\textbf{0.719}), with a small difference to the optimized default settings (\textbf{0.706}). Default \acrshort{hp} settings from different \acrshort{ml} tools presented a very similar distribution and the same performance (\textbf{0.644}), which is worse than that of \texttt{default.opt}.


The Friedman test was applied to assess the statistical significance of the \acrshort{hp} strategies considering a significance level of $\alpha = 0.05$~\citep{SantaFe:2015}. The null hypothesis states that all the strategies have equivalent predictive performance.
When the null hypothesis is rejected, the Nemenyi post-hoc test is also used to indicate which strategies are significantly different.

Figure~\ref{fig:nemenyi_samples-51} presents the resultant \acrfull{cd} diagram. In this diagram, strategies are connected when there are \textit{no} significant differences between them.
Therefore, according to this test, there is no significant difference between  \acrshort{rs} and the new optimized settings (default.opt). It is important to highlight that, in spite of its simplicity, \acrshort{rs} optimizes the \acrshort{hp} values for each dataset, evaluating a high number of candidate solutions. Thus, its computational cost is much higher than the cost of evaluating the pool of the settings obtained by the proposed strategy, \texttt{default.opt}.
In summary, these results show that default.opt is able to find appropriate \acrshort{hp} settings since it outperformed the \acrshort{ml} tools and was competitive with \acrshort{rs}.


\begin{figure}[h!]
    \centering
    \includegraphics
    [scale = 0.35]
    {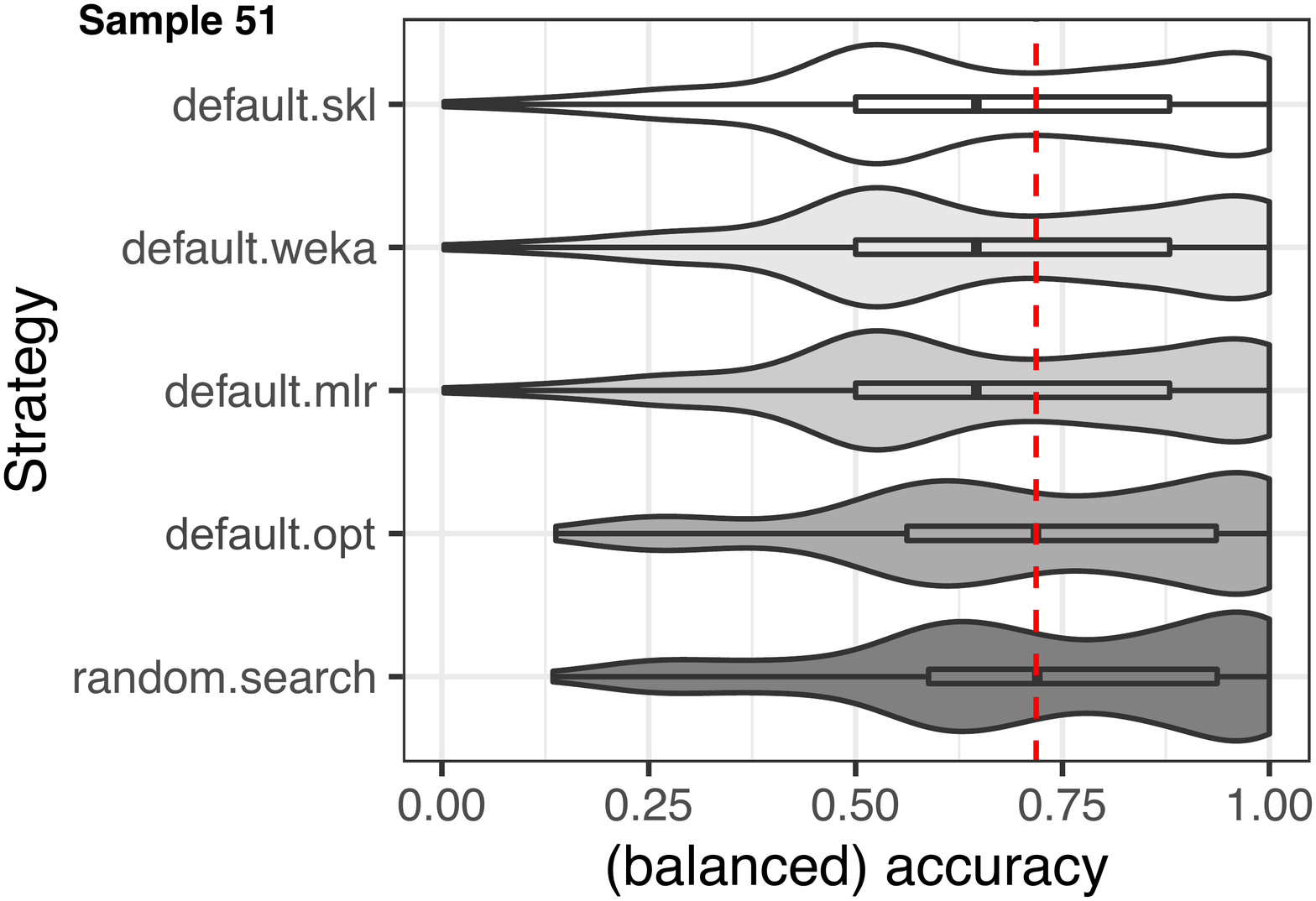}
    \caption{\acrshort{bac} performance distributions obtained by different strategies evaluated in the test datasets and default.opt with $k=51$.}
    \label{fig:violin_performance_sample_51}

    \vspace{0.3cm}
    \centering
    
    
        \begin{tikzpicture}[xscale=1.5]
            \node (Label) at (1.3860480558075452, 0.7){\tiny{CD = 0.31}}; 
            \draw[decorate,decoration={snake,amplitude=.4mm,segment length=1.5mm,post length=0mm},very thick, color = black] (1.2,0.5) -- (1.5720961116150902,0.5);
            \foreach \x in {1.2, 1.5720961116150902} \draw[thick,color = black] (\x, 0.4) -- (\x, 0.6);
             
            \draw[gray, thick](1.2,0) -- (6.0,0); 
            \foreach \x in {1.2,2.4,3.6,4.8,6.0} \draw (\x cm,1.5pt) -- (\x cm, -1.5pt);
            \node (Label) at (1.2,0.2){\tiny{1}};
            \node (Label) at (2.4,0.2){\tiny{2}};
            \node (Label) at (3.6,0.2){\tiny{3}};
            \node (Label) at (4.8,0.2){\tiny{4}};
            \node (Label) at (6.0,0.2){\tiny{5}};
            \draw[decorate,decoration={snake,amplitude=.4mm,segment length=1.5mm,post length=0mm},very thick, color = black](2.2135658914728684,-0.25) -- (2.5833333333333335,-0.25);
            \draw[decorate,decoration={snake,amplitude=.4mm,segment length=1.5mm,post length=0mm},very thick, color = black](4.337596899224806,-0.25) -- (4.47015503875969,-0.25);
            \node (Point) at (2.2635658914728682, 0){};\node (Label) at (0.5,-0.65){\scriptsize{random.search}}; \draw (Point) |- (Label);
            \node (Point) at (2.5333333333333337, 0){};\node (Label) at (0.5,-0.95){\scriptsize{default.opt}}; \draw (Point) |- (Label);
            \node (Point) at (4.42015503875969, 0){};\node (Label) at (6.5,-0.65){\scriptsize{default.mlr}}; \draw (Point) |- (Label);
            \node (Point) at (4.395348837209302, 0){};\node (Label) at (6.5,-0.95){\scriptsize{default.skl}}; \draw (Point) |- (Label);
            \node (Point) at (4.387596899224806, 0){};\node (Label) at (6.5,-1.25){\scriptsize{default.weka}}; \draw (Point) |- (Label);

    \end{tikzpicture}
    \label{fig:sample_51_cd_005}
     
 
    \caption[The CD diagram considering the \acrshort{bac} values of the HP settings found by different strategies (default.opt with $k=51$) for \acrshortpl{svm} according to the Frideman-Nemenyi test ($\alpha = 0.05$).]{The CD diagram considering the \acrshort{bac} values of the HP settings found by different strategies (default.opt with $k=51$) for \acrshortpl{svm} according to the Frideman-Nemenyi test ($\alpha = 0.05$). Groups of strategies that are not significantly different are connected.}
    
    \label{fig:cd_diagram_tree_51}
    \label{fig:nemenyi_samples-51}
  
\end{figure}


\subsection{Improvement Analysis}
\label{sec:improvement-analysis}

For a more in-depth analysis of the improvements achieved by our strategy, Figure~\ref{fig:svm_agg_improvements} shows the \acrshort{bac} values (y-axis) for each dataset (x-axis) averaged over the 5x2 \acrshort{cv} resampling results.
Different colors and shapes of lines represent different \acrshort{hp} strategies.
These datasets are named by their \texttt{OpenML} IDs and are listed by decreasing \acrshort{bac} values obtained by \texttt{default.mlr} (LibSVM). 

\afterpage{%
    \rotatebox{90}{
    \begin{minipage}{0.8\paperheight}
        \includegraphics[width=\textwidth,keepaspectratio]
        {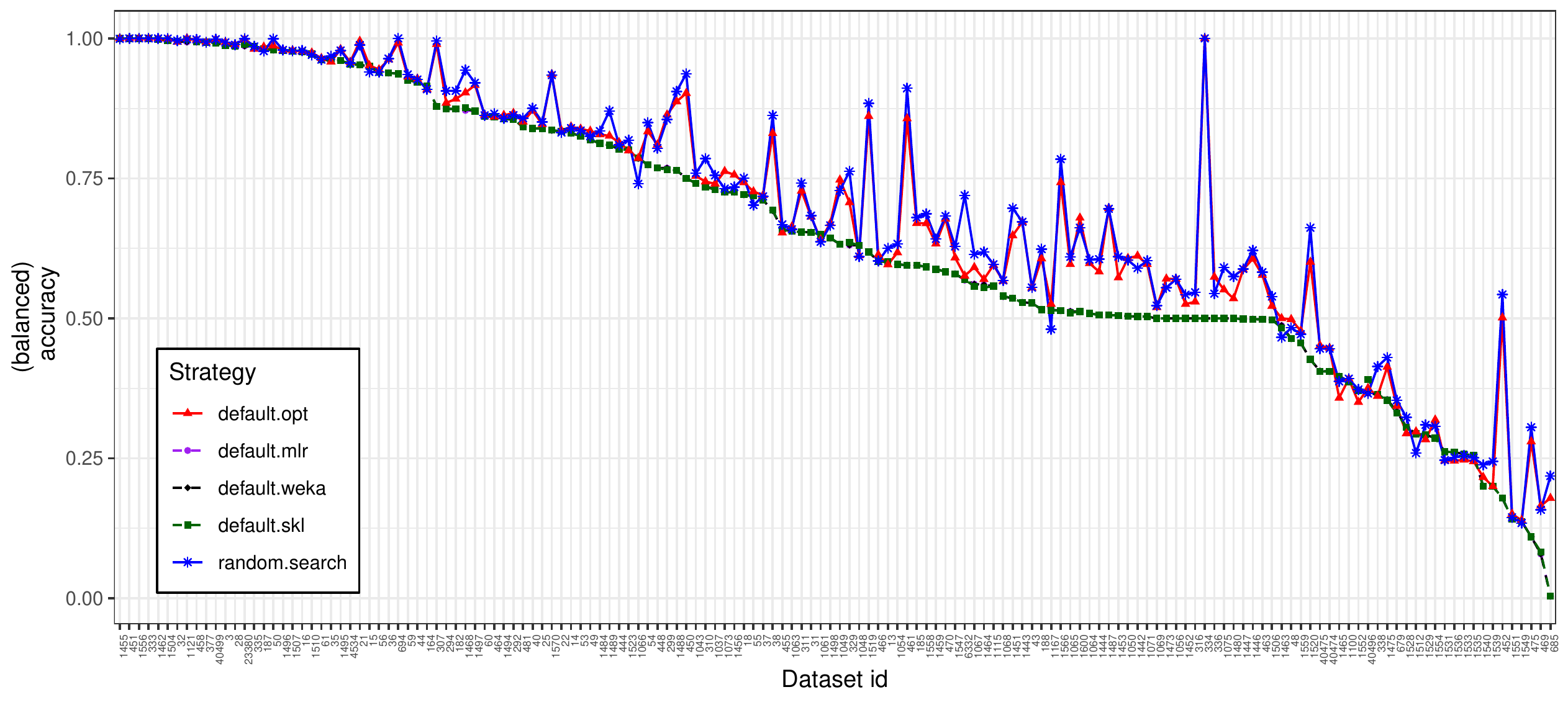}
        \captionof{figure}{\acrshort{hp} tuning results in \acrshortpl{svm} in different test sets.}
        \label{fig:svm_agg_improvements}
    \end{minipage}
    }
}

Figure~\ref{fig:svm_agg_improvements} shows that default \acrshort{hp} settings from \acrshort{ml} tools (mlr, Weka and scikit-learn) performed similarly for all  datasets, what was expected given the overall analysis, and, thus, their curves were usually overlapped.
The new optimized \acrshort{hp} settings (red line), generated by our strategy, and the \acrshort{rs} technique (blue line) outperformed them in most of the datasets. This figure also shows that the behavior of the default.opt's curve is very similar to the \acrshort{rs} curve, including the cases of performance gains.
Our initial hypothesis was that \acrshort{rs} would defeat the new optimized default settings. 
Thus, this achievement is surprising to some extent given that \acrshort{rs} is performing \acrshort{hp} tuning for each dataset, and is consequently much more time consuming when compared to the few evaluations needed to evaluate a small set of optimized settings.


The Wilcoxon paired-test (with $\alpha=0.05$) was applied to assess the statistical significance of the results between the two best-ranked \acrshort{hp} strategies per dataset considering the averaged \acrshort{bac} values presented in Figure~\ref{fig:svm_agg_improvements}.
Table~\ref{tab:wilcoxon_aggregated} presents the frequency each strategy was best ranked with (p-value $< 0.05$) and without (p-value $\geq 0.05$) statistical significance when compared to the second best strategy. The comparison \textit{RS vs Tools} only considers defaults from the \acrshort{ml} tools whereas \textit{RS vs All} also considers the optimized defaults. 


\begin{table}[ht!]
    \centering
    \caption{Wilcoxon paired test comparing the two best \acrshort{hp} strategies ranked by data set. For each \acrshort{hp} strategy, the frequency with which it was ranked first with (p-value $<$ 0.05) and without (p-value $\geq$ 0.05) statistical significance is presented. Table presents aggregated results for all the different test sets used.}
    
    \label{tab:wilcoxon_aggregated}
    \begin{tabular}{lrrrr}
    \toprule
  
    \multirow{2}{*}{\textbf{Strategy}} & \multicolumn{2}{c}{\textbf{RS vs Tools}} & \multicolumn{2}{c}{\textbf{RS vs All}} \\
    & $<0.05$ & $\geq0.05$ & $<0.05$ & $\geq0.05$\\
   
    \midrule 
 
    \textbf{Random Search (RS)} & 89 & 37 & 38 & 45 \\
    \textbf{Default opt} & -- & -- & 10 & 43\\
    \textbf{Default mlr} & 0 & 15 & 0 & 10\\
    \textbf{Default skl}  & 0 & 9 & 0 & 4\\
    \textbf{Default Weka} & 0 & 0 & 0 & 0 \\
    
    \midrule 
    \multicolumn{1}{l}{Total of test cases} & \multicolumn{2}{r}{150} & \multicolumn{2}{r}{150} 
    \\ 
    \bottomrule
    \end{tabular}
\end{table}


Overall, the \acrshort{rs} technique significantly outperformed ($< 0.05$)  default settings from tools in $89$ datasets. For the remaining ($ \geq 0.05$), there was no significant difference among \acrshort{rs} and tools, but the former was ranked as the first strategy for $37$ out of $61$ datasets.  Therefore, \acrshort{hp} tuning is  recommended for these $89$ datasets, since in practice, due to its computational cost, we would employ it just when it is likely to significantly improve predictive performance. Some studies, such as \cite{Mantovani:2019}, showed that it is possible to predict with high accuracy whether or not a process of \acrshort{hp} tuning is necessary for the dataset under analysis. 
When the default optimized settings are included in this comparison (\textit{RS vs All}), they were able to remarkably reduce the number of datasets where \acrshort{rs} is recommended to $38$ ($< 0.05$). Besides, optimized defaults were significantly better than the second ranked strategy for $10$ cases. Thus, using the method proposed by~\cite{Mantovani:2019}, the \acrshort{hp} tuning would be performed on only $38$ datasets, instead of $89$.


\begin{figure}[htbp]
    \centering
    \includegraphics
    {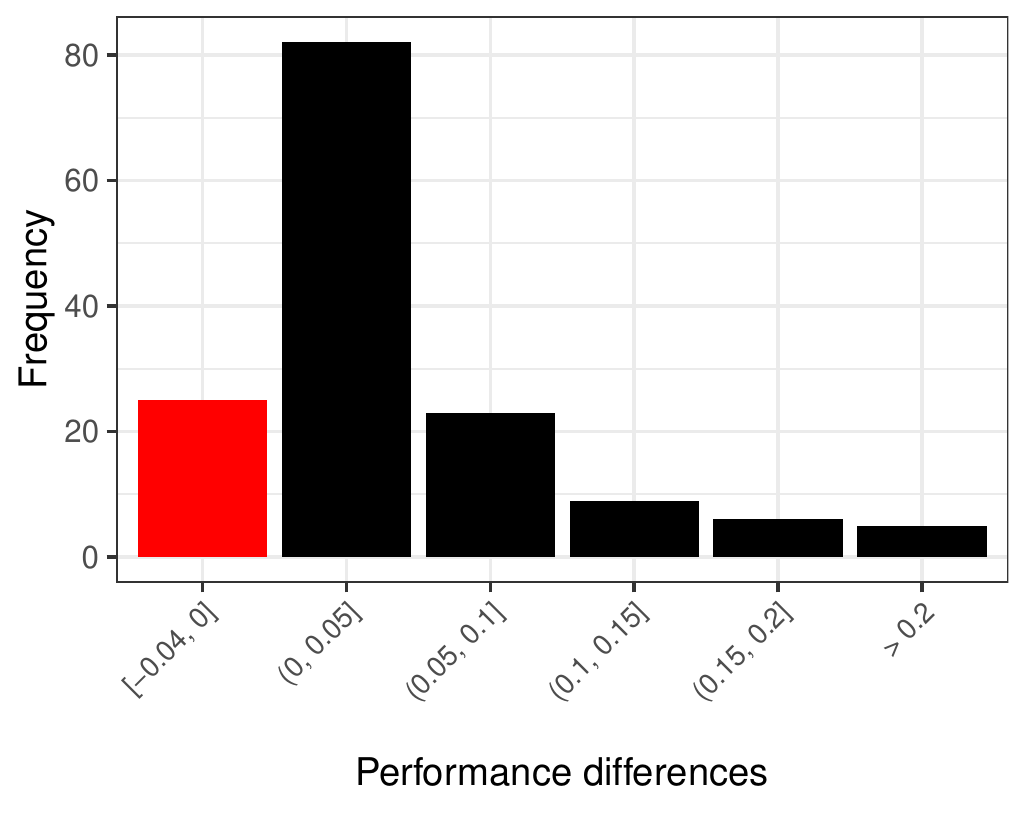}
    \caption{Performance differences in terms of \acrshort{bac} when comparing results obtained by the default optimized \acrshort{hp} settings to the defaults from \acrshort{ml} tools.}
    \label{fig:svm_histograms}
\end{figure}



In order to compare the pool of optimized default \acrshort{hp} settings with those provided by the \acrshort{ml} tools, Figure~\ref{fig:svm_histograms} shows the distribution of the predictive performance difference assessed by \acrshort{bac} values for all datasets.
Black bars represent favorable differences to our strategy (positive differences indicate improvement) whereas red bars indicate when traditional defaults were better (negative differences). 
Observing the red bar, one can notice the traditional defaults only achieved minimal advantages since they are mostly close to zero. On the other hand, when using the new optimized \acrshort{hp} settings, there are cases with medium (between $0.05$ and $0.1$) and high improvement (above $0.1$).


\subsection{Analysis of the new optimized \acrshort{svm} \acrshort{hp} values}
\label{sec:analsyis_def_opt}

\begin{figure}[h!]
        \centering
        \includegraphics[scale=0.55]
        {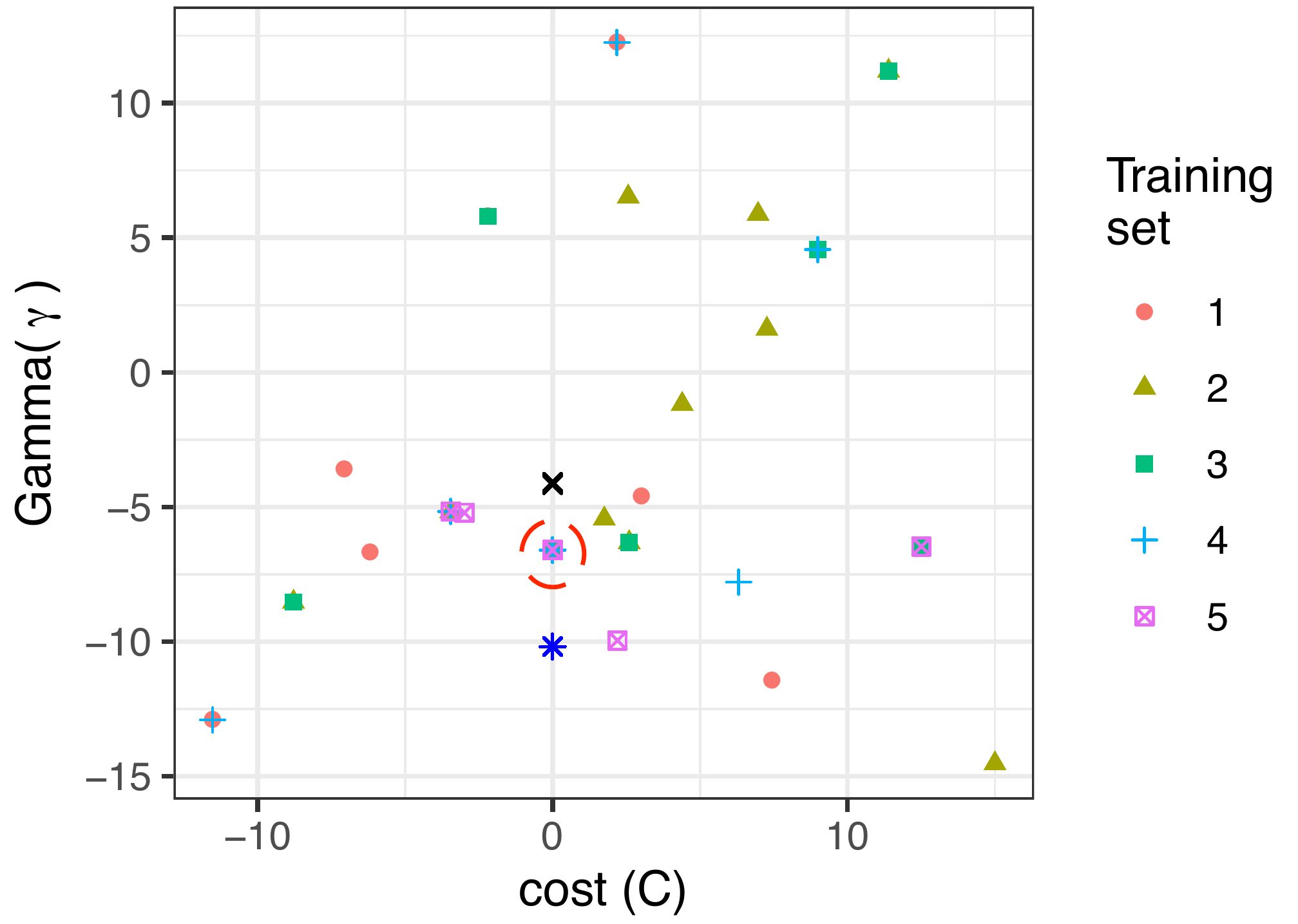}
        \captionof{figure}{The dispersion of the new optimized \acrshortpl{hp} settings obtained with sample size $k=51$. Values projected on the $log_2$ scale. The black cross and blue point shows default values from \acrshort{ml} tools in datasets where our strategy performed best and worst, respectively.}
        \label{fig:hp_dispersion}
\end{figure}

Figure~\ref{fig:hp_dispersion} depicts the dispersion of the new optimized \acrshort{hp} settings in the \acrshort{svm} \acrshort{hp} space. The x-axis shows the cost (C) values and the y-axis shows the gamma ($\gamma$) values, both in the $log_{2}$ scale.
Different shapes and colors denote \acrshort{hp} settings obtained considering the five different $D_{opt}$ sets. The dashed circle indicates the region containing the \acrshort{hp} values included in the initial population of the optimization method. \\
In addition, the black cross and the blue point represent default values from \acrshort{ml} tools in datasets where our strategy performed best\footnote{https://www.openml.org/d/334} and worst\footnote{https://www.openml.org/d/4550}, respectively. 


\begin{table}[h!]
        \centering
        \captionof{table}{Top-ranked \acrshort{hp} settings obtained with shared optimization and sample size = 51. Values are represented in $log_2$ scale.}
        \label{tab:hp_dispersion}
        \begin{tabular}{crr}
        \toprule
        \multirow{2}{*}{\textbf{Rank}} & \multicolumn{2}{c}{\textbf{HP Setting}} \\
        & \textbf{Cost (C)} & \textbf{Gamma ($\gamma$)}\\
         \midrule
            1  & -2.1927  & 5.7930    \\
            2  & 3.0154   & -4.5968   \\
            3  & 8.9897   & 4.5561    \\
            4  & 0.0000   & -6.6000   \\
            5  & 12.5062  & -6.4680   \\
            6  & 7.4370   & -11.4271  \\ 
            7  & -7.0694  & -3.5971   \\
            8  & -6.1878  & -6.6787   \\
            9  & -11.5290 &  -12.9075 \\
            10 & 2.1856  & 12.2462    \\
            \bottomrule
          \end{tabular}
\end{table}


According to this figure, there is a dispersion of the new optimized \acrshort{hp} settings across the hyperspace. Besides, different training sets influenced the optimization process differently, with their \acrshort{hp} settings located in different regions:
\begin{itemize}
   
    \item Training sets 1, 2, 3, 4: these were able to generate various \acrshort{hp} settings across the investigated datasets. Their \acrshort{hp} values differ from traditional defaults. 
    Thus, the search explored different regions from the space, and these different settings were able to induce models with good performance, often better than traditional defaults;
    
    \item Training set 5: provided few variability on their optimized settings, with most of them placed near the initial search space. Thus, one may argue that optimization became stuck in a local minimum, considering those datasets, and did not explore the remainder of the space. Nevertheless, even with values closer to the traditional \acrshort{ml} defaults, they were competitive to the \acrshort{rs} baseline, as shown in Figure~\ref{fig:svm_agg_improvements} and Table~\ref{tab:wilcoxon_aggregated}.
\end{itemize}

The top-ranked \acrshort{hp} settings obtained by the optimization average over all test sets ($D_{test}$) are presented in Table~\ref{tab:hp_dispersion}.
Appendix~\ref{app:defaults} presents an extended table with all the unique \acrshort{hp} settings we obtained in our experiments with the best experimental setup (sample size $k = 51$).


\subsection{Learning from new optimized defaults}
\label{sec:learning_from_def_opt}

Although the strategy to find optimized default settings works well, some questions regarding its use may arise.
For example, ``what both dataset and learning characteristics can tell us about when to use the pool of optimized \acrshort{hp} settings instead of a tuning technique, such as \acrshort{rs}?''.
The answer to this question can help users to choose between either \acrshort{hp} tuning, which can provide better settings but is computationally expensive, or testing a set of default \acrshort{hp} values. Moreover, finding some patterns regarding this task may bring some knowledge about the learning process.

Here, we borrow some ideas from \acrfull{mtl}~\cite{Lemke:2015} to investigate this problem as a binary classification task where classes identify whether the pool of default optimized \acrshort{hp} settings are sufficient for a given dataset or \acrshort{hp} tuning using \acrshort{rs} should be performed.
\acrshort{hp} tuning is recommended when it significantly outperformed the new default \acrshort{hp} settings for a given dataset based on the Wilcoxon test, previously discussed in Section~\ref{sec:improvement-analysis}.
The predictive attributes consist of different characteristics extracted from each dataset using the \texttt{pymfe}\footnote{https://github.com/ealcobaca/pymfe}(\textit{v0.4}) tool, with the mean (\texttt{nanmean}) and standard deviation (\texttt{nanstd}) summary functions~\citep{Alcobaca:2020}. 


\begin{figure}
    \centering
    \includegraphics
    [width=0.9\textwidth]
    {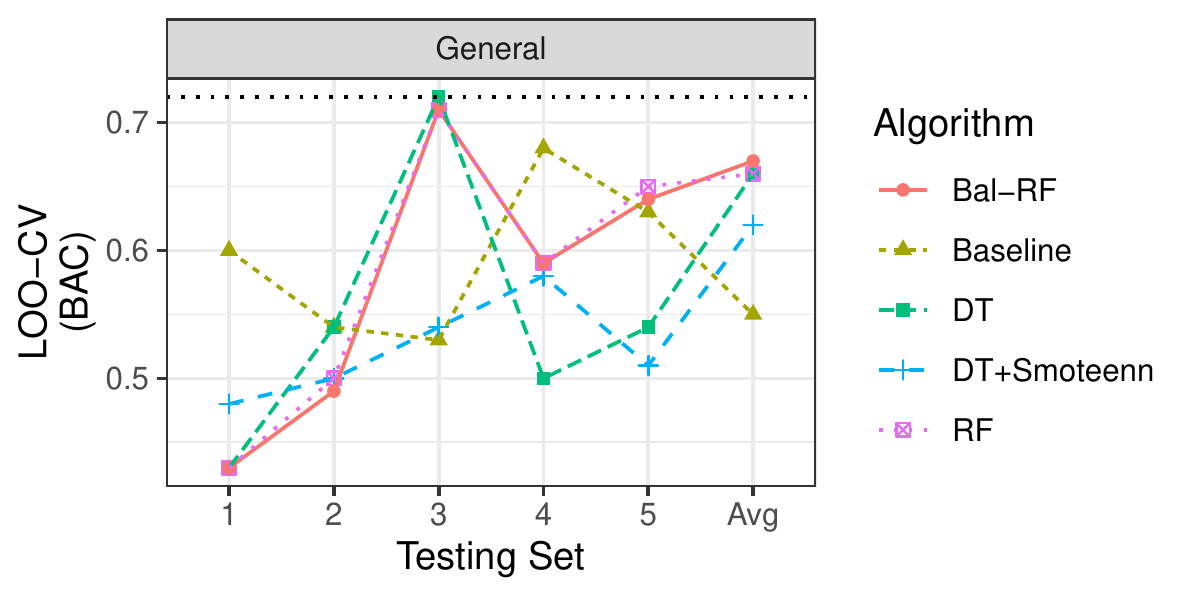}
    \caption{\acrshort{bac} of the LOO-CV using simple and general meta-feature groups with the sample size $k=51$.}
    \label{fig:meta-scenarios-result}
\end{figure}


We considered in our analysis each test set, and the average of these sets for the sample size $k=51$.
The data generated for each one was used by the \acrfull{dt} and \acrfull{rf} algorithms to induce predictive models.
These algorithms were selected because they can induce explainable models, and are therefore able to shed some light on the learning process.
Due to the presence of class imbalance in the meta-dataset, we also included an \acrshort{rf} version that deals with imbalance (Bal-RF), a \acrshort{dt} performing oversampling via SMOTE, and cleaning using ENN (Smoteenn). These \acrshort{ml} and preprocessing algorithms are available in the \texttt{scikit-learn} and \texttt{imbalanced-learn} Python libraries~\citep{Lematre:2017, Pedregosa:2015}. Their results were compared with a baseline model that always predicts the majority class.

Figure~\ref{fig:meta-scenarios-result} shows the \acrshort{bac} of the \acrfull{loocv} (y-axis) considering  only simple and general descriptors for each test set and their average.
In this figure, different algorithms are represented by different colors, line types and shapes. 
Overall, the best result was obtained by the \acrshort{dt} algorithm with a \acrshort{bac} of $0.72$ for the third test set, which is $0.19$ higher than the baseline, suggesting that this algorithm could learn some patterns from these data.
On the other hand, the induced models did not overcome the baseline for some test sets, which may be an indicator that we have a difficult learning problem.


\begin{figure}[phtb] 
\centering 
        \includegraphics
        [scale=1]
        {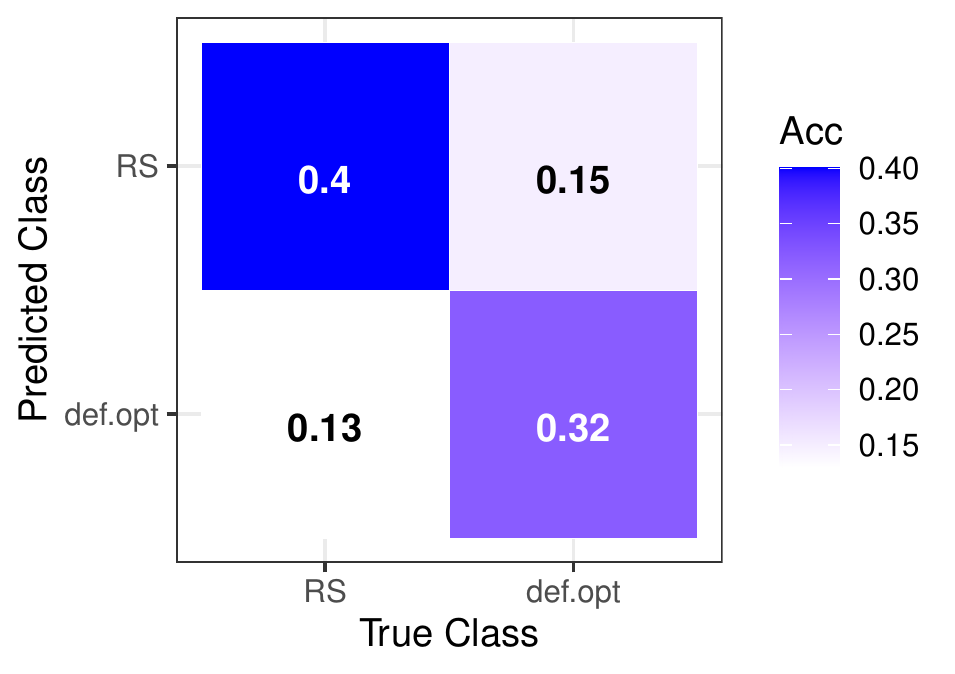}
        \caption{Confusion matrix of the best induced \acrshort{dt} model (0.72) in test set 3, with general and simple descriptors and not using the Smoteenn.}
        \label{fig:conf_matrix_best_dt}
\end{figure}


Figure~\ref{fig:conf_matrix_best_dt} shows the confusion matrix for the best \acrshort{dt} model. 
The difference between the accuracies for each class indicates that this model is more prone to hit the \acrshort{rs} class than the def.opt class. 
This behavior is desired when predictive performance is the main concern, since it spends more computational time performing hp tuning but avoid performance loss.

Although we have shown that these models are better than a guess, we have no access to which characteristics make a dataset more prone to \acrshort{rs} than to the pool of optimized \acrshort{hp} yet. Therefore, we draw the best \acrshort{dt} model to understand the learned patterns. Due to its size, the complete \acrshort{dt} model is presented in Appendix~\ref{app:figures}. There, we present the entire \acrshort{dt} model, its branches, leaves, the yielded rules, and a histogram of the features.

Based on this model, we can observe two interesting rule paths where more than half of the dataset is included. 
The first rule is shown next, where $20$ out of the $23$ examples (datasets) are from the \texttt{default.opt} class and the remaining $3$ examples are from the \texttt{rs} class. This rule evaluates the number of instances from a dataset (\texttt{nr\_inst}), the total number of attributes (\texttt{nr\_attr}) and the proportion of these two characteristics, i.e., the number of attributes divided by the number of instances  (\texttt{attr\_to\_instance}).
This rule suggests that optimized \acrshort{hp} settings are more suitable for datasets with a small number of examples ($<358$) and attributes ($<88$). It may be motivated by the nature of most of datasets presented in the \acrshort{uci} and \acrshort{openml} repositories, mostly ''simple' problems.


\begin{Verbatim}[frame=single]
nr_inst < 358
  attr_to_instance >= 0.02
      nr_attr < 88
          [20/3] (default.opt/RS)
\end{Verbatim}


The second interesting rule path has $28$ examples, $24$ from
the \texttt{RS} class and $4$ from the \texttt{default.opt} class. This rule suggests that the tuning process should be performed for datasets with more examples (\texttt{nr\_inst} $\geq 358$) and balanced classes (with a standard deviation of the relative frequency of each class $ \geq 0.43$). 
Overall, the induced tree presents interesting patterns by using simple dataset characteristics. 
Moreover, as this tree is small, the practitioners can use it to identify when to use default settings or perform \acrshort{hp} tuning for their new problems.

\begin{Verbatim}[frame=single]
nr_inst >= 358
  freq_class.nansd >= 0.43
      freq_class.nanmean >= 0.15
          [4/24] (default.opt/RS)
\end{Verbatim}


\subsection{Sample Size Sensitivity Analysis}
\label{sec:sample_size_sensitivity}

In the previous sections, we analyzed how suitable are the \acrshort{hp} settings found by the proposed strategy considering the best sample size studied ($k=51$). 
In this section, we analyze how this strategy behaves for different sample sizes, i.e., we investigate the effect of the sample size in the \acrshort{hp} settings starting with few datasets and increasing it until almost all training datasets ($D_{train}$) are used: $k={11,31,51,71}$.

\begin{figure}[h]
    
    \centering
    \hspace{1.8cm}
    \includegraphics[scale = 1]
    {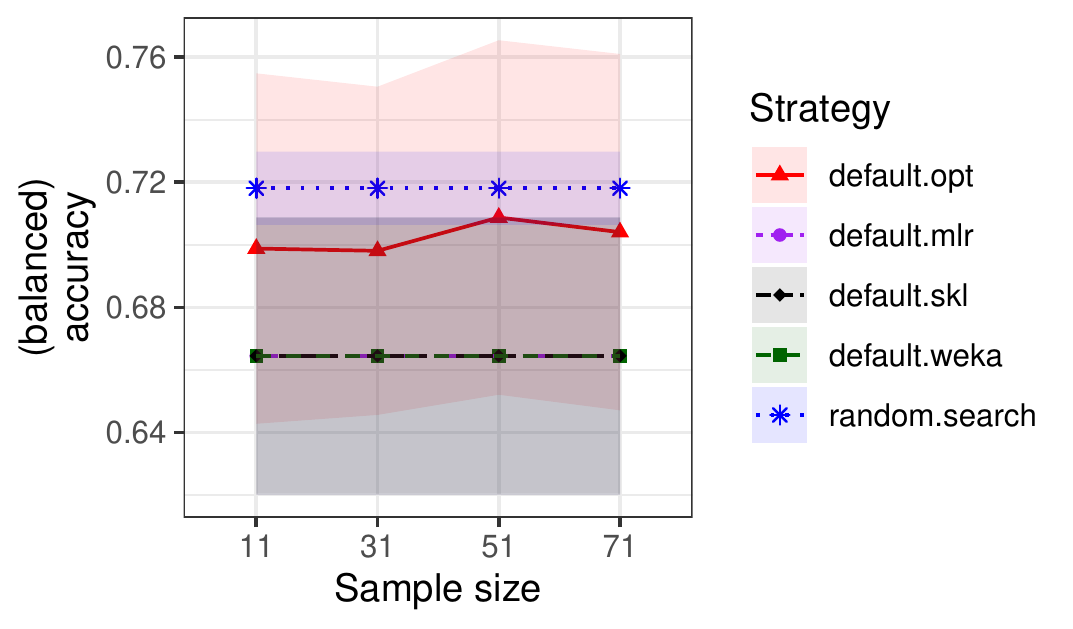}
    
    \caption{Mean \acrshort{bac} performance values evaluated in test datasets with different sample sizes.}
    \label{fig:avg_performance_sample}

    \vspace{0.4cm}

    \centering
    \begin{tikzpicture}[xscale=1.5]
        \node (Label) at (1.678817186646894, 0.7){\tiny{CD = 0.24}}; 
        \draw[decorate,decoration={snake,amplitude=.4mm,segment length=1.5mm,post length=0mm},very thick, color = black] (1.5,0.5) -- (1.8576343732937879,0.5);
        \foreach \x in {1.5, 1.8576343732937879} \draw[thick,color = black] (\x, 0.4) -- (\x, 0.6);
         
        \draw[gray, thick](1.5,0) -- (6.0,0); 
        \foreach \x in {1.5,3.0,4.5,6.0} \draw (\x cm,1.5pt) -- (\x cm, -1.5pt);
        \node (Label) at (1.5,0.2){\tiny{1}};
        \node (Label) at (3.0,0.2){\tiny{2}};
        \node (Label) at (4.5,0.2){\tiny{3}};
        \node (Label) at (6.0,0.2){\tiny{4}};
        \draw[decorate,decoration={snake,amplitude=.4mm,segment length=1.5mm,post length=0mm},very thick, color = black](3.513953488372093,-0.25) -- (3.902713178294573,-0.25);
        \draw[decorate,decoration={snake,amplitude=.4mm,segment length=1.5mm,post length=0mm},very thick, color = black](3.8027131782945736,-0.4) -- (4.038372093023256,-0.4);
        \node (Point) at (3.5639534883720927, 0){};\node (Label) at (0.5,-0.65){\scriptsize{Sample Size = 51}}; \draw (Point) |- (Label);
        \node (Point) at (3.5949612403100772, 0){};\node (Label) at (0.5,-0.95){\scriptsize{Sample Size = 71}}; \draw (Point) |- (Label);
        \node (Point) at (3.988372093023256, 0){};\node (Label) at (6.5,-0.65){\scriptsize{Sample Size = 11}}; \draw (Point) |- (Label);
        \node (Point) at (3.8527131782945734, 0){};\node (Label) at (6.5,-0.95){\scriptsize{Sample Size = 31}}; \draw (Point) |- (Label);
    \end{tikzpicture}
    \caption{Comparison of the \acrshort{bac} values of the different \acrshort{hp} settings according to the Friedman-Nemenyi test ($\alpha = 0.05)$ for different sample sizes.}
    \label{fig:nemenyi_sizes}
    
\end{figure}

Figure~\ref{fig:avg_performance_sample} shows the average BAC achieved by default.opt and the baselines. This figure suggests that our strategy is benefiting from bigger sample sizes, i.e., $k=\{51, 71\}$ datasets,  when there is a clear approximation to the \acrshort{rs} and a larger distance to the defaults of \acrshort{ml} tools. Even with the average values attenuating their differences and with a high standard deviation (shaded colors), both cases show a promising scenario.


To confirm whether these sample sizes yielded different performances, we also applied the Friedman test with a significance level $\alpha = 0.05$. 
The null hypothesis states that all the sample sizes are equivalent with respect to the \acrfull{bac} values.
Figure~\ref{fig:nemenyi_samples} presents the resultant \acrfull{cd} diagram. The smallest sample ($k=11$) was statistically worse than sample sizes $k=\{51, 71\}$. In all the remaining cases, there is no significant differences despite the bigger difference between $k=\{31\}$ and $k=\{51,71\}$.
This analysis suggests that smaller samples are less appropriate. However, although we did not have statistical evidence to choose a sample size among $k=\{31,51,71\}$, we chose the best one ($k=\{51\}$) for the previous experimental analysis.

In the proposed strategy, the fitness value to find default optimized settings  considers the predictive performance across different datasets (see Fig.~\ref{fig:exp-strategy}). Thus, one could argue that the higher the number of datasets used in this process the better the \acrshort{hp} settings found, and, consequently, the predictive performance. This probably would be true if we had considered only the best setting. Instead, we are interested in analyzing a pool of settings. Thus, there is a trade-off between the sample size and the diversity, i.e., smaller sample sizes are more likely to generate settings that cover different regions of the search space.



\section{Threats to Validity} 
\label{sec:validity}

In an empirical study design, methodological choices may impact the results obtained in the experiments. Next, we discuss the threats that may impact the results of this study.



\subsection{Internal validity}



The datasets used in the experiments were preprocessed to be handled by \acrshortpl{svm}. We also ensured that all classes in the datasets must have at least $10$ observations. 
Thus, 10-fold stratification can be applied without any concerns.
Of course, other datasets may be used to expand data collection, if they comply with the `stratified' criterion. However, the authors believe that adding datasets will not substantially change the overall behavior of \acrshort{hp} strategies on the algorithm investigated, since they were selected to cover a wide range of classification tasks with different characteristics.


\cite{Krstajic:2014} compared different resampling strategies for assessing the predictive performance and selecting  regression/classification models induced by \acrshort{ml} algorithms. In~\cite{Cawley:2010}, the authors also discuss the overfitting in the evaluation methodologies when assessing \acrshort{ml} algorithms.
Based on their discussion, the most reasonable choice for our experiments is the 5 times 2-CV resampling methodology. It is suggested for cases when it is desired to reduce the variance of the results generated using a dataset with few instances. It is exactly our case: we have a few instances ($150$) which are datasets feeding an optimization process. 
Thus, to reduce the bias of the performance evaluation and the computational cost of experiments, this resampling methodology was adopted in the experiments.


Since a wide variety of datasets compose the data collection, some of them may be imbalanced. Thus, the \acrshort{bac} measure~\citep{Brodersen:2010} was used to assess the predictive performance of the models during the optimization process, i.e., in the fitness function. This measure considers class distributions when assessing the performance of a candidate solution. We used the same performance measure to evaluate the final solutions returned by the \acrshort{hp} strategies. Other predictive performance measures can generate different results, depending on how they deal with data imbalance.


\subsection{Conclusion validity}



Section~\ref{sec:results} presented statistical comparisons between the investigated tuning strategies. In~\cite{Demvsar:2006}, the author discusses the issue of statistical tests for comparisons of several techniques on multiple datasets, reviewing several statistical methodologies. The method proposed as more suitable is the non-parametric analog version of ANOVA, i.e., the Friedman test, along with the corresponding Nemenyi post-hoc test. The Friedman test ranks all the methods separately for each dataset and uses the average ranks to test whether all techniques are equivalent. In case of differences, the Nemenyi test performs all the pairwise comparisons between the techniques and identifies the presence of significant differences. Thus, the Friedman ranking test followed by the Nemenyi post-hoc test was used to evaluate our experimental results.


\subsection{External validity}


The experimental methodology described in Section~\ref{sec:meth} considers using \acrshort{hp} tuning techniques that have been used and discussed in the literature, such as~\cite{Pfisterer:2018, Rijn:2018}. The \acrshort{pso} and \acrshort{rs} techniques were also exhaustively benchmarked for \acrshort{svm} tuning~\citep{Mantovani:2018-thesis}. 
In the experiments carried out in this paper, we used the default settings provided by the \acrshort{pso} R implementation. These default values are robust for our dataset collection. Otherwise, 
the tuning of \acrshort{pso} would considerably increase the experimental cost by adding a new level of tuning (\textit{the tuning of tuning techniques}). Thus, this additional level was not assessed in this study.


Using budgets for \acrshortpl{svm} tuning was investigated in~\cite{mantovani:2015-rs}. The experimental results suggested that all the considered techniques required only $\approx$ $300$ evaluations to converge. Convergence here means the tuning techniques could not improve their predictive performance more than $10^{-5}$ until the budget was consumed. Actually, in most cases, the tuning reached its maximum performance after $100$ steps. Thus, a budget size of $300$ evaluations was therefore deemed sufficient. Results obtained from this budget showed that the exploration made in hyperparameter spaces led to statistically significant improvements in most cases. Thus, this budget size was adopted in our experiments.


In this paper, we investigated a single \acrshort{ml} algorithm. The methodology described here can be generalized to other \acrshort{ml} algorithms, especially those that are sensitive to tuning. On the other hand, algorithms such as the \acrshort{rf} whose defaults are robust enough~\citep{probst2018} would not benefit from the new strategy. 
Nonetheless, additional similar studies may prove fruitful. For such, all the experimental data generated in the experiments are available at \texttt{OpenML}\footnote{{\url{https://www.openml.org/s/52}}} and \texttt{GitHub}\footnote{\url{https://github.com/rgmantovani/OptimDefaults}}.


\section{Conclusion}
\label{sec:conc}

The present paper proposes and rigorously analyzes a strategy to generate  optimized \acrshort{hp} settings for \acrshort{ml} algorithms. 
This strategy was conceived from the observation that a limited number of default \acrshort{hp} values is provided by most of the \acrshort{ml} tools. Therefore, alternative default settings could improve model predictive performance besides reducing the need to perform \acrshort{hp} tuning, which is a very time-consuming task.

For such, we carried out experiments adopting \acrfullpl{svm} as the \acrshort{ml} algorithm due to its sensitivity to \acrshort{hp} tuning and used a collection of $150$ datasets carefully curated from the public repository \acrshort{openml}.
In addition, the \acrshort{pso} optimization technique was used to find \acrshort{hp} settings that are appropriate for a sample of this collection of datasets. 

Using this new set of \acrshort{hp} values, referred to as default optimized, led to significantly better models than the defaults suggested by \acrshort{ml} tools in all scenarios investigated.
Furthermore, the optimized default settings were able to considerably reduce the number of datasets for which it is necessary to perform \acrshort{hp} tuning.
A sensitivity analysis of the strategy regarding sample size suggested there is a trade-off between finding robust \acrshort{hp} settings, what requires a high number of datasets, and promoting diversity, using a smaller sample.
In our experiments, the best results were obtained with a sample of $51$ datasets. 

The ideal situation would be that where we could define for which problems default settings are sufficient and for which ones the \acrshort{hp} tuning is indicated.
Thus, we conducted experiments using characteristics extracted from the dataset collection and tree-based algorithms to provide some interpretability of the identified patterns. In our analysis, two main rules could be observed: (i) default optimized settings is a better option when a dataset has a small number of examples and attributes; 
and (ii) \acrshort{hp} tuning is usually recommended for datasets with more examples and balanced classes. The robustness of these new optimized \acrshort{hp} settings goes toward what has been discussed in the literature~\citep{Bergstra:2012, Mantovani:2019, Weerts:2020}.


\subsection{Main difficulties}

The main difficulties faced during this study are related to the cost of the performed experiments.
The optimization process is computationally expensive, since a large number of datasets are evaluated for every new candidate solution (\acrshort{hp} setting). We also needed to run several rounds of experiments to calibrate our methodology, and each of these rounds took almost two months.

Another adversity is related to the data collection. Initially, a larger number of datasets were selected, but some of them presented problems when inducing \acrshortpl{svm} models. 
Therefore, we preferred to not consider them in the experiments.


\subsection{Future Work}

The findings from this study open up future research directions.
In the context of \acrshort{automl}, the obtained \acrshort{hp} settings can be used as a warm start for the optimization techniques.
Moreover, instead of creating pipelines from scratch, the \acrshort{automl} systems can create entire pipelines based only on the optimized defaults. 

It would also be a promising direction to investigate different ways of coding the individuals in the optimization process. The sample size and correspondent datasets could be embedded in the candidate solution, along with the \acrshort{hp} values, releasing designers from these empirical choices.

Another possibility would be to cluster the datasets according to their similarities to generate better-optimized \acrshort{hp} values for each group. The fitness value used in the experiments is an aggregate measure of performance across different datasets. It would be interesting to explore other measures, such as average ranks.

The code used in this study is publicly available, easily extendable, and may be adapted to cover several other \acrshort{ml} algorithms. Thus, experiments with different \acrshort{ml} algorithms can also be carried out, investigating their \acrshort{hp} profile: the need for tuning, how defaults behave, and so on. All the information generated can also be used as meta-knowledge to feed further experiments.


\section*{Acknowledgments}

\begin{sloppypar}
The authors would like to thank CAPES and CNPq, grants 420562/2018-4 and 309863/2020-1, (Brazilian Agencies) for their financial support, especially for grants \#2012/23114-9, \#2015/03986-0, \#2016/18615-0 and \#2018/14819-5 from the S{\~a}o Paulo Research Foundation (FAPESP). This research was also carried out using the computational resources of the Center for Mathematical Sciences Applied to Industry (CeMEAI) funded by FAPESP (grant \#2013/07375-0).
\end{sloppypar}



\appendix
\newpage
\clearpage
\section{List of abbreviations used in the paper} 
\label{app:glossary}
\printnoidxglossary[type=\acronymtype, title=\empty]


\newpage
\section{List of datasets used in the experiments} 
\label{app:datasets}

\begin{table}[ht!]
\scriptsize
\setlength{\tabcolsep}{2pt}
\centering
 \caption{(Multi-class) classification OpenML datasets (1 to 43) used in experiments. It is shown, for each dataset: the OpenML dataset name and id, the number of attributes (D), the number of examples (N), the number of classes (C), the number of examples belonging to the majority and minority classes (nMaj, nMin), and the proportion between them (P).}

\begin{tabular}{clcrrrrrc}
  \toprule
  
  \textbf{Nro } & \textbf{ OpenML name} & \textbf{OpenML did} & \textbf{ D } & \textbf{ N } & \textbf{ C } & \textbf{ nMaj } & \textbf{ nMin } & \textbf{ P} \\ 
  
  \midrule
  
  1 & kr-vs-kp &   3 &  37 & 3196 &   2 & 1669 & 1527 & 0.91 \\ 
  2 & breast-cancer &  13 &  10 & 286 &   2 & 201 &  85 & 0.42 \\ 
  3 & mfeat-fourier &  14 &  77 & 2000 &  10 & 200 & 200 & 1.00 \\ 
  4 & breast-w &  15 &  10 & 699 &   2 & 458 & 241 & 0.53 \\ 
  5 & mfeat-karhunen &  16 &  65 & 2000 &  10 & 200 & 200 & 1.00 \\ 
  6 & mfeat-morphological &  18 &   7 & 2000 &  10 & 200 & 200 & 1.00 \\ 
  7 & car &  21 &   7 & 1728 &   4 & 1210 &  65 & 0.05 \\ 
  8 & mfeat-zernike &  22 &  48 & 2000 &  10 & 200 & 200 & 1.00 \\ 
  9 & colic &  25 &  28 & 368 &   2 & 232 & 136 & 0.59 \\ 
  10 & optdigits &  28 &  65 & 5620 &  10 & 572 & 554 & 0.97 \\ 
  11 & credit-g &  31 &  21 & 1000 &   2 & 700 & 300 & 0.43 \\ 
  12 & pendigits &  32 &  17 & 10992 &  10 & 1144 & 1055 & 0.92 \\ 
  13 & dermatology &  35 &  35 & 366 &   6 & 112 &  20 & 0.18 \\ 
  14 & segment &  36 &  20 & 2310 &   7 & 330 & 330 & 1.00 \\ 
  15 & diabetes &  37 &   9 & 768 &   2 & 500 & 268 & 0.54 \\ 
  16 & sick &  38 &  30 & 3772 &   2 & 3541 & 231 & 0.07 \\ 
  17 & sonar &  40 &  61 & 208 &   2 & 111 &  97 & 0.87 \\ 
  18 & haberman &  43 &   4 & 306 &   2 & 225 &  81 & 0.36 \\
  19 & spambase &  44 &  58 & 4601 &   2 & 2788 & 1813 & 0.65 \\ 
  20 & tae &  48 &   6 & 151 &   3 &  52 &  49 & 0.94 \\ 
  21 & heart-c &  49 &  14 & 303 &   5 & 165 &   0 & 0.00 \\
  22 & tic-tac-toe &  50 &  10 & 958 &   2 & 626 & 332 & 0.53 \\ 
  23 & heart-statlog &  53 &  14 & 270 &   2 & 150 & 120 & 0.80 \\ 
  24 & vehicle &  54 &  19 & 846 &   4 & 218 & 199 & 0.91 \\ 
  25 & hepatitis &  55 &  20 & 155 &   2 & 123 &  32 & 0.26 \\ 
  26 & vote &  56 &  17 & 435 &   2 & 267 & 168 & 0.63 \\ 
  27 & ionosphere &  59 &  35 & 351 &   2 & 225 & 126 & 0.56 \\ 
  28 & waveform-5000 &  60 &  41 & 5000 &   3 & 1692 & 1653 & 0.98 \\ 
  29 & iris &  61 &   5 & 150 &   3 &  50 &  50 & 1.00 \\ 
  30 & molecular-biology\_promoters & 164 &  59 & 106 &   2 &  53 &  53 & 1.00 \\ 
  31 & satimage & 182 &  37 & 6430 &   6 & 1531 & 625 & 0.41 \\ 
  32 & baseball & 185 &  18 & 1340 &   3 & 1215 &  57 & 0.05 \\ 
  33 & wine & 187 &  14 & 178 &   3 &  71 &  48 & 0.68 \\ 
  34 & eucalyptus & 188 &  20 & 736 &   5 & 214 & 105 & 0.49 \\ 
  35 & Australian & 292 &  15 & 690 &   2 & 383 & 307 & 0.80 \\ 
  36 & satellite\_image & 294 &  37 & 6435 &   6 & 871 & 275 & 0.32 \\ 
  37 & libras\_move & 299 &  91 & 360 &  11 &  24 &  11 & 0.46 \\ 
  38 & vowel & 307 &  13 & 990 &  11 &  90 &  90 & 1.00 \\ 
  39 & mammography & 310 &   7 & 11183 &   2 & 10923 & 260 & 0.02 \\ 
  40 & oil\_spill & 311 &  50 & 937 &   2 & 896 &  41 & 0.05 \\ 
  41 & yeast\_ml8 & 316 & 117 & 2417 &   2 & 2383 &  34 & 0.01 \\ 
  42 & hayes-roth & 329 &   5 & 160 &   4 &  65 &   0 & 0.00 \\ 
  43 & monks-problems-1 & 333 &   7 & 556 &   2 & 278 & 278 & 1.00 \\ 

   \bottomrule
\end{tabular}
\label{tab:data1}
\end{table}

\clearpage

\begin{table}[ht]
\scriptsize
\setlength{\tabcolsep}{2pt}
\centering
 \caption{(Multi-class) classification OpenML datasets (44 to 88) used in experiments. For each dataset it is shown: the OpenML dataset name and id, the number of attributes (D), the number of examples (N), the number of classes (C), the number of examples belonging to the majority and minority classes (nMaj, nMin), and the proportion between them (P).}

\begin{tabular}{clcrrrrrc}
  \toprule
  
  \textbf{Nro } & \textbf{ OpenML name} & \textbf{OpenML did} & \textbf{ D } & \textbf{ N } & \textbf{ C } & \textbf{ nMaj } & \textbf{ nMin } & \textbf{ P} \\ 
  
  \midrule
   44 & monks-problems-2 & 334 &   7 & 601 &   2 & 395 & 206 & 0.52 \\ 
   45 & monks-problems-3 & 335 &   7 & 554 &   2 & 288 & 266 & 0.92 \\ 
   46 & SPECT & 336 &  23 & 267 &   2 & 212 &  55 & 0.26 \\ 
  47 & grub-damage & 338 &   9 & 155 &   4 &  49 &  19 & 0.39 \\ 
  48 & synthetic\_control & 377 &  62 & 600 &   6 & 100 & 100 & 1.00 \\ 
  49 & analcatdata\_boxing2 & 444 &   4 & 132 &   2 &  71 &  61 & 0.86 \\ 
  50 & analcatdata\_boxing1 & 448 &   4 & 120 &   2 &  78 &  42 & 0.54 \\ 
  51 & analcatdata\_lawsuit & 450 &   5 & 264 &   2 & 245 &  19 & 0.08 \\ 
  52 & irish & 451 &   6 & 500 &   2 & 278 & 222 & 0.80 \\ 
  53 & analcatdata\_broadwaymult & 452 &   8 & 285 &   7 & 118 &  21 & 0.18 \\ 
  54 & cars & 455 &   9 & 406 &   3 & 254 &  73 & 0.29 \\ 
  55 & analcatdata\_authorship & 458 &  71 & 841 &   4 & 317 &  55 & 0.17 \\ 
  56 & analcatdata\_creditscore & 461 &   7 & 100 &   2 &  73 &  27 & 0.37 \\ 
  57 & backache & 463 &  33 & 180 &   2 & 155 &  25 & 0.16 \\ 
  58 & prnn\_synth & 464 &   3 & 250 &   2 & 125 & 125 & 1.00 \\ 
  59 & schizo & 466 &  15 & 340 &   2 & 177 & 163 & 0.92 \\ 
  60 & analcatdata\_dmft & 469 &   5 & 797 &   6 & 155 & 123 & 0.79 \\ 
  61 & profb & 470 &  10 & 672 &   2 & 448 & 224 & 0.50 \\ 
  62 & analcatdata\_germangss & 475 &   6 & 400 &   4 & 100 & 100 & 1.00 \\ 
  63 & biomed & 481 &   9 & 209 &   2 & 134 &  75 & 0.56 \\ 
  64 & rmftsa\_sleepdata & 679 &   3 & 1024 &   4 & 404 &  94 & 0.23 \\ 
  65 & visualizing\_livestock & 685 &   3 & 130 &   5 &  26 &  26 & 1.00 \\ 
  66 & diggle\_table\_a2 & 694 &   9 & 310 &   9 &  41 &  18 & 0.44 \\ 
  67 & ada\_prior & 1037 &  15 & 4562 &   2 & 3430 & 1132 & 0.33 \\ 
  68 & ada\_agnostic & 1043 &  49 & 4562 &   2 & 3430 & 1132 & 0.33 \\ 
  69 & jEdit\_4.2\_4.3 & 1048 &   9 & 369 &   2 & 204 & 165 & 0.81 \\ 
  70 & pc4 & 1049 &  38 & 1458 &   2 & 1280 & 178 & 0.14 \\ 
  71 & pc3 & 1050 &  38 & 1563 &   2 & 1403 & 160 & 0.11 \\ 
  72 & mc2 & 1054 &  40 & 161 &   2 & 109 &  52 & 0.48 \\ 
  73 & mc1 & 1056 &  39 & 9466 &   2 & 9398 &  68 & 0.01 \\ 
  74 & ar4 & 1061 &  30 & 107 &   2 &  87 &  20 & 0.23 \\ 
  75 & kc2 & 1063 &  22 & 522 &   2 & 415 & 107 & 0.26 \\ 
  76 & ar6 & 1064 &  30 & 101 &   2 &  86 &  15 & 0.17 \\ 
  77 & kc3 & 1065 &  40 & 458 &   2 & 415 &  43 & 0.10 \\ 
  78 & kc1-binary & 1066 &  95 & 145 &   2 &  85 &  60 & 0.71 \\ 
  79 & kc1 & 1067 &  22 & 2109 &   2 & 1783 & 326 & 0.18 \\ 
  80 & pc1 & 1068 &  22 & 1109 &   2 & 1032 &  77 & 0.07 \\ 
  81 & pc2 & 1069 &  37 & 5589 &   2 & 5566 &  23 & 0.00 \\ 
  82 & mw1 & 1071 &  38 & 403 &   2 & 372 &  31 & 0.08 \\ 
  83 & jEdit\_4.0\_4.2 & 1073 &   9 & 274 &   2 & 140 & 134 & 0.96 \\ 
  84 & datatrieve & 1075 &   9 & 130 &   2 & 119 &  11 & 0.09 \\ 
  85 & PopularKids & 1100 &  11 & 478 &   3 & 247 &  90 & 0.36 \\ 
  86 & teachingAssistant & 1115 &   7 & 151 &   3 &  52 &  49 & 0.94 \\ 
  87 & badges2 & 1121 &  12 & 294 &   2 & 210 &  84 & 0.40 \\ 
  88 & pc1\_req & 1167 &   9 & 320 &   2 & 213 & 107 & 0.50 \\ 
  
   \bottomrule
\end{tabular}
\label{tab:data2}
\end{table}


\begin{table}[ht]
\scriptsize
\setlength{\tabcolsep}{2pt}
\centering
 \caption{(Multi-class) classification OpenML datasets (89 to 134) used in experiments. For each dataset it is shown: the OpenML dataset name and id, the number of attributes (D), the number of examples (N), the number of classes (C), the number of examples belonging to the majority and minority classes (nMaj, nMin), and the proportion between them (P).}

\begin{tabular}{clcrrrrrc}
  \toprule
  
  \textbf{Nro } & \textbf{ OpenML name} & \textbf{OpenML did} & \textbf{ D } & \textbf{ N } & \textbf{ C } & \textbf{ nMaj } & \textbf{ nMin } & \textbf{ P} \\ 
  
  \midrule

  89 & MegaWatt1 & 1442 &  38 & 253 &   2 & 226 &  27 & 0.12 \\ 
  90 & PizzaCutter1 & 1443 &  38 & 661 &   2 & 609 &  52 & 0.09 \\ 
  91 & PizzaCutter3 & 1444 &  38 & 1043 &   2 & 916 & 127 & 0.14 \\ 
  92 & CostaMadre1 & 1446 &  38 & 296 &   2 & 258 &  38 & 0.15 \\ 
  93 & CastMetal1 & 1447 &  38 & 327 &   2 & 285 &  42 & 0.15 \\ 
  94 & PieChart1 & 1451 &  38 & 705 &   2 & 644 &  61 & 0.09 \\ 
  95 & PieChart2 & 1452 &  37 & 745 &   2 & 729 &  16 & 0.02 \\ 
  96 & PieChart3 & 1453 &  38 & 1077 &   2 & 943 & 134 & 0.14 \\ 
  97 & acute-inflammations & 1455 &   7 & 120 &   2 &  70 &  50 & 0.71 \\ 
  98 & appendicitis & 1456 &   8 & 106 &   2 &  85 &  21 & 0.25 \\ 
  99 & artificial-characters & 1459 &   8 & 10218 &  10 & 1416 & 600 & 0.42 \\ 
  100 & banknote-authentication & 1462 &   5 & 1372 &   2 & 762 & 610 & 0.80 \\

  101 & blogger & 1463 &   6 & 100 &   2 &  68 &  32 & 0.47 \\ 
  102 & blood-transfusion-service-center & 1464 &   5 & 748 &   2 & 570 & 178 & 0.31 \\ 
  103 & breast-tissue & 1465 &  10 & 106 &   6 &  22 &  14 & 0.64 \\ 
  104 & cnae-9 & 1468 & 857 & 1080 &   9 & 120 & 120 & 1.00 \\ 
  105 & fertility & 1473 &  10 & 100 &   2 &  88 &  12 & 0.14 \\ 
  106 & first-order-theorem-proving & 1475 &  52 & 6118 &   6 & 2554 & 486 & 0.19 \\ 
  107 & ilpd & 1480 &  11 & 583 &   2 & 416 & 167 & 0.40 \\ 
  108 & lsvt & 1484 & 311 & 126 &   2 &  84 &  42 & 0.50 \\ 
  109 & ozone-level-8hr & 1487 &  73 & 2534 &   2 & 2374 & 160 & 0.07 \\ 
  110 & parkinsons & 1488 &  23 & 195 &   2 & 147 &  48 & 0.33 \\ 
  111 & phoneme & 1489 &   6 & 5404 &   2 & 3818 & 1586 & 0.42 \\ 
  112 & qsar-biodeg & 1494 &  42 & 1055 &   2 & 699 & 356 & 0.51 \\ 
  113 & qualitative-bankruptcy & 1495 &   7 & 250 &   2 & 143 & 107 & 0.75 \\ 
  114 & ringnorm & 1496 &  21 & 7400 &   2 & 3736 & 3664 & 0.98 \\ 
  115 & wall-robot-navigation & 1497 &  25 & 5456 &   4 & 2205 & 328 & 0.15 \\ 
  116 & sa-heart & 1498 &  10 & 462 &   2 & 302 & 160 & 0.53 \\ 
  117 & steel-plates-fault & 1504 &  34 & 1941 &   2 & 1268 & 673 & 0.53 \\ 
  118 & thoracic-surgery & 1506 &  17 & 470 &   2 & 400 &  70 & 0.18 \\ 
  119 & twonorm & 1507 &  21 & 7400 &   2 & 3703 & 3697 & 1.00 \\ 
  120 & wdbc & 1510 &  31 & 569 &   2 & 357 & 212 & 0.59 \\ 
  121 & heart-long-beach & 1512 &  14 & 200 &   5 &  56 &  10 & 0.18 \\ 
  122 & robot-failures-lp4 & 1519 &  91 & 117 &   3 &  72 &  21 & 0.29 \\ 
  123 & robot-failures-lp5 & 1520 &  91 & 164 &   5 &  47 &  21 & 0.45 \\ 
  124 & vertebra-column & 1523 &   7 & 310 &   3 & 150 &  60 & 0.40 \\ 
  125 & volcanoes-a2 & 1528 &   4 & 1623 &   5 & 1471 &  29 & 0.02 \\ 
  126 & volcanoes-a3 & 1529 &   4 & 1521 &   5 & 1369 &  29 & 0.02 \\ 
  127 & volcanoes-b1 & 1531 &   4 & 10176 &   5 & 9791 &  26 & 0.00 \\ 
  128 & volcanoes-b3 & 1533 &   4 & 10386 &   5 & 10006 &  25 & 0.00 \\ 
  129 & volcanoes-b5 & 1535 &   4 & 9989 &   5 & 9599 &  26 & 0.00 \\ 
  130 & volcanoes-b6 & 1536 &   4 & 10130 &   5 & 9746 &  26 & 0.00 \\ 
  131 & volcanoes-d2 & 1539 &   4 & 9172 &   5 & 8670 &  56 & 0.01 \\ 
  132 & volcanoes-d3 & 1540 &   4 & 9285 &   5 & 8771 &  58 & 0.01 \\ 
  133 & autoUniv-au1-1000 & 1547 &  21 & 1000 &   2 & 741 & 259 & 0.35 \\ 
  134 & autoUniv-au6-750 & 1549 &  41 & 750 &   8 & 165 &  57 & 0.35 \\
 
   \bottomrule
\end{tabular}
\label{tab:data3}
\end{table}

\begin{table}[ht]
\scriptsize
\setlength{\tabcolsep}{2pt}
\centering
 \caption{(Multi-class) classification OpenML datasets (135 to 150) used in experiments. For each dataset it is shown: the OpenML dataset name and id, the number of attributes (D), the number of examples (N), the number of classes (C), the number of examples belonging to the majority and minority classes (nMaj, nMin), and the proportion between them (P).}

\begin{tabular}{clcrrrrrc}
  \toprule
  
  \textbf{Nro } & \textbf{ OpenML name} & \textbf{OpenML did} & \textbf{ D } & \textbf{ N } & \textbf{ C } & \textbf{ nMaj } & \textbf{ nMin } & \textbf{ P} \\ 
  
  \midrule

 135 & autoUniv-au6-400 & 1551 &  41 & 400 &   8 & 111 &  25 & 0.23 \\ 
  136 & autoUniv-au7-1100 & 1552 &  13 & 1100 &   5 & 305 & 153 & 0.50 \\ 
  137 & autoUniv-au7-500 & 1554 &  13 & 500 &   5 & 192 &  43 & 0.22 \\ 
  138 & acute-inflammations & 1556 &   7 & 120 &   2 &  61 &  59 & 0.97 \\ 
  139 & bank-marketing & 1558 &  17 & 4521 &   2 & 4000 & 521 & 0.13 \\ 
  140 & breast-tissue & 1559 &  10 & 106 &   4 &  49 &  14 & 0.29 \\ 
  141 & hill-valley & 1566 & 101 & 1212 &   2 & 612 & 600 & 0.98 \\ 
  142 & wilt & 1570 &   6 & 4839 &   2 & 4578 & 261 & 0.06 \\ 
  143 & SPECTF & 1600 &  45 & 267 &   2 & 212 &  55 & 0.26 \\ 
  144 & PhishingWebsites & 4534 &  31 & 11055 &   2 & 6157 & 4898 & 0.80 \\ 
  145 & cylinder-bands & 6332 &  40 & 540 &   2 & 312 & 228 & 0.73 \\
  146 & cjs & 23380 & 36 & 2796 & 6 & 576 & 341 & 0.59 \\
  147 & thyroid-allbp & 40474 &  27 & 2800 &   5 & 1632 &  31 & 0.02 \\ 
  148 & thyroid-allhyper & 40475 &  27 & 2800 &   5 & 1632 &  31 & 0.02 \\ 
  149 & LED-display-domain-7digit & 40496 &   8 & 500 &  10 &  57 &  37 & 0.65 \\ 
  150 & texture & 40499 &  41 & 5500 &  11 & 500 & 500 & 1.00 \\ 
   \bottomrule
\end{tabular}
\label{tab:data4}
\end{table}


\clearpage
\newpage
\section{List of HP settings generated in the experiments defaults.} 
\label{app:defaults}

\begin{table*}[th!]
        \centering
        \caption{HP settings obtained with shared optimization and sample size = 51. Values are represented in $log_2$ scale. HP settings with a bullet ($\bullet$) were obtained by more than one training/test resamplings.}
        \label{tab:shared_hps}
        \setlength{\tabcolsep}{9pt}
        \begin{tabular}{crrccc}
        \toprule
        \multirow{2}{*}{\textbf{Rank}} & \multicolumn{2}{c}{\textbf{HP Setting}} & \textbf{Fitness} & \multirow{2}{*}{\textbf{Test set}} & \multirow{2}{*}{\textbf{+}}\\
        & \textbf{Cost (C)} & \textbf{Gamma ($\gamma$)} & \textbf{Value} & & \\
         \midrule
             1  & -2.192770    &   5.793062  & 0.6987194 & 1 & $\bullet$ \\    
             2  &    3.015420  &  -4.596853  & 0.6965455 & 1 \\ 
             3  &    8.989739  &   4.556140  & 0.6965080 & 1 & $\bullet$ \\  
             4  &    0.000000  &  -6.600000  & 0.6944090 & 1 & $\bullet$ \\  
             5  &   12.506273  &  -6.468016  & 0.6941373 & 1 & $\bullet$ \\  
             6  &   7.437093   & -11.427108  & 0.6921446 & 1 \\  
             7  &   -7.069438  &  -3.597182  & 0.6920009 & 1 \\   
             8  &   -6.187812  &  -6.678751  & 0.6919634 & 1 \\  
             9  &  -11.529067  & -12.907540  & 0.6916917 & 1 & $\bullet$ \\   
             10  &   2.185601  &  12.246234  & 0.6899332 & 1 & $\bullet$ \\  
             11  &   6.311317  &  -7.790445  & 0.6666667 & 4 \\   
             12  &  11.391777  &  11.190030  & 0.6603367 & 3 & $\bullet$ \\   
             13  &  -3.451729  &  -5.167970  & 0.6585796 & 5 & $\bullet$ \\   
             14  &   2.597285  &  -6.316462  & 0.6585185 & 3 & $\bullet$ \\   
             15  &   2.199790  &  -9.958442  & 0.6580640 & 5 \\   
             16  &  -8.786429  &  -8.527994  & 0.6576431 & 3 & $\bullet$ \\   
             17  &  -2.989745  &  -5.215827  & 0.6562795 & 5 \\   
             18  &   2.565695  &   6.517172  & 0.6057916 & 2 \\   
             19  &   4.393857  &  -1.182761  & 0.6004798 & 2 \\   
             20  &   6.967503  &   5.874307  & 0.5992790 & 2 \\  
             21  &   1.751779  &  -5.436291  & 0.5981402 & 2 \\  
             22  &  15.000000  & -14.527203  & 0.5978719 & 2 \\  
             23  &   7.267743  &   1.608208  & 0.5952172 & 2 \\  
        \bottomrule
          \end{tabular}
\end{table*}


\newpage
\section{List of Meta-features used in experiments} 
\label{app:features}

\begin{table*}[h!]
\centering
\caption{Meta-features used in experiments. For each meta-feature it is shown: its type, acronym and description. Adapted from~\texttt{PyMFE} documentation*.}
\label{tab:mfeats_1}
\begin{tabular}{lll}
  
  \toprule
  \multicolumn{1}{l}{\textbf{Type}} & \textbf{Acronym} & \textbf{Description} \\
  \midrule
  
  \multirow{11}{*}{General}
  & attr\_to\_inst & Compute the ratio between the number of attributes \\
  & cat\_to\_num & Compute the ratio between the number of categorical and numerical features \\
  & freq\_class & Compute the relative frequency of each distinct class \\
  & inst\_to\_attr & Compute the ratio between the number of instances and attributes \\
  & nr\_attr & Compute the total number of attributes \\
  & nr\_bin & Compute the number of binary attributes \\
  & nr\_cat & Compute the number of categorical attributes \\
  & nr\_class & Compute the number of distinct classes \\
  & nr\_inst & Compute the number of instances (rows) in the dataset \\
  & nr\_num & Compute the number of numeric features \\
  & num\_to\_cat & Compute the number of numerical and categorical features. \\
  \bottomrule

\end{tabular}
\\
* - \url{https://pymfe.readthedocs.io/en/latest/auto_examples/index.html}
\end{table*}


\section{Decision tree model obtained during the experiments} 
\label{app:figures}

\rotatebox{90}{
\begin{minipage}{\textheight}
    \includegraphics[width=\textwidth,keepaspectratio]{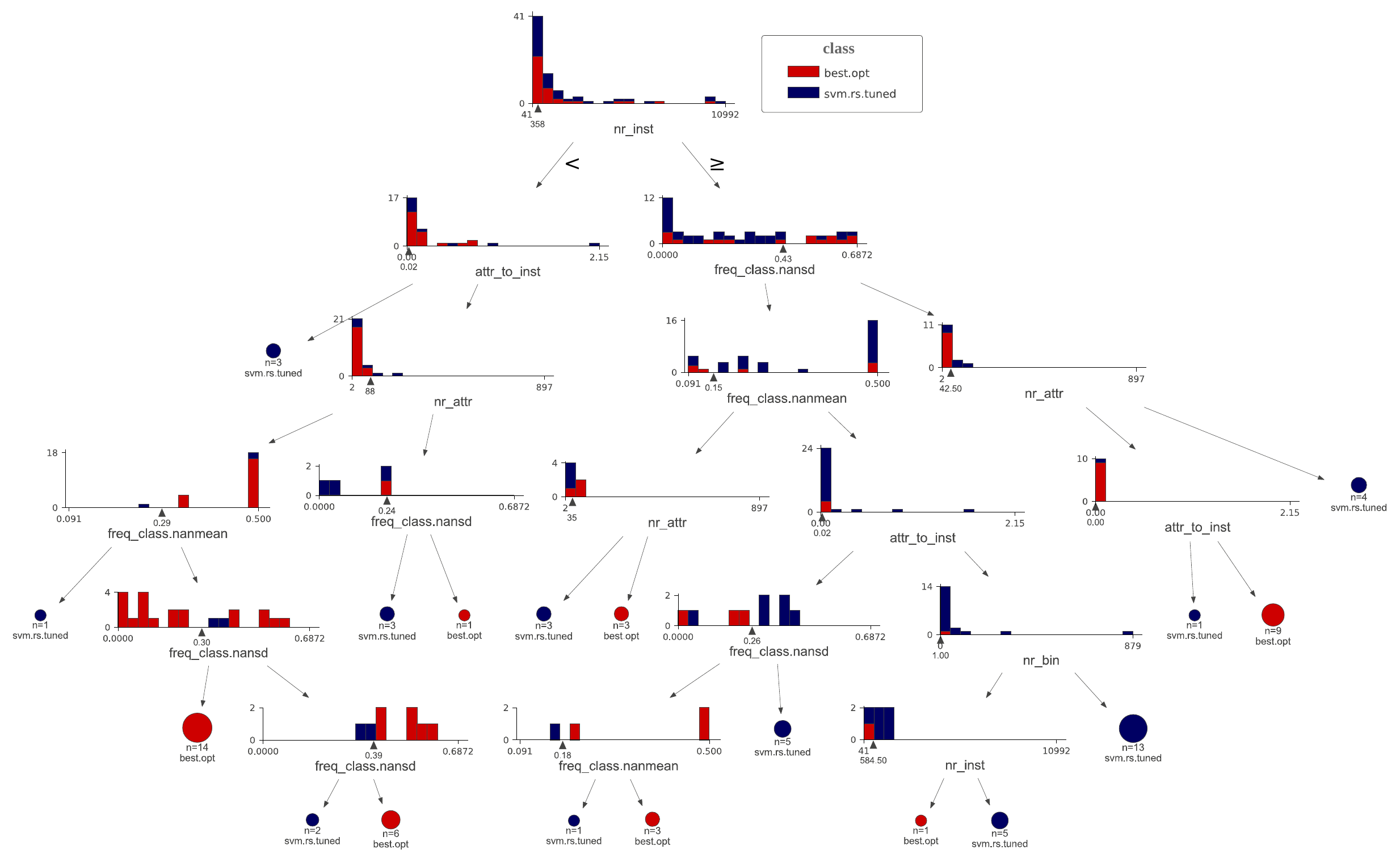}
    \captionof{figure}{The decision rules of the best DT model found.}
    \label{fig:dt_tree}
\end{minipage}
}


\newpage


\section{Results of all sample sizes} 
\label{sec:all-sample-sizes}

\begin{figure}[h!]
    \centering
    \includegraphics
    [width=\textwidth]
    {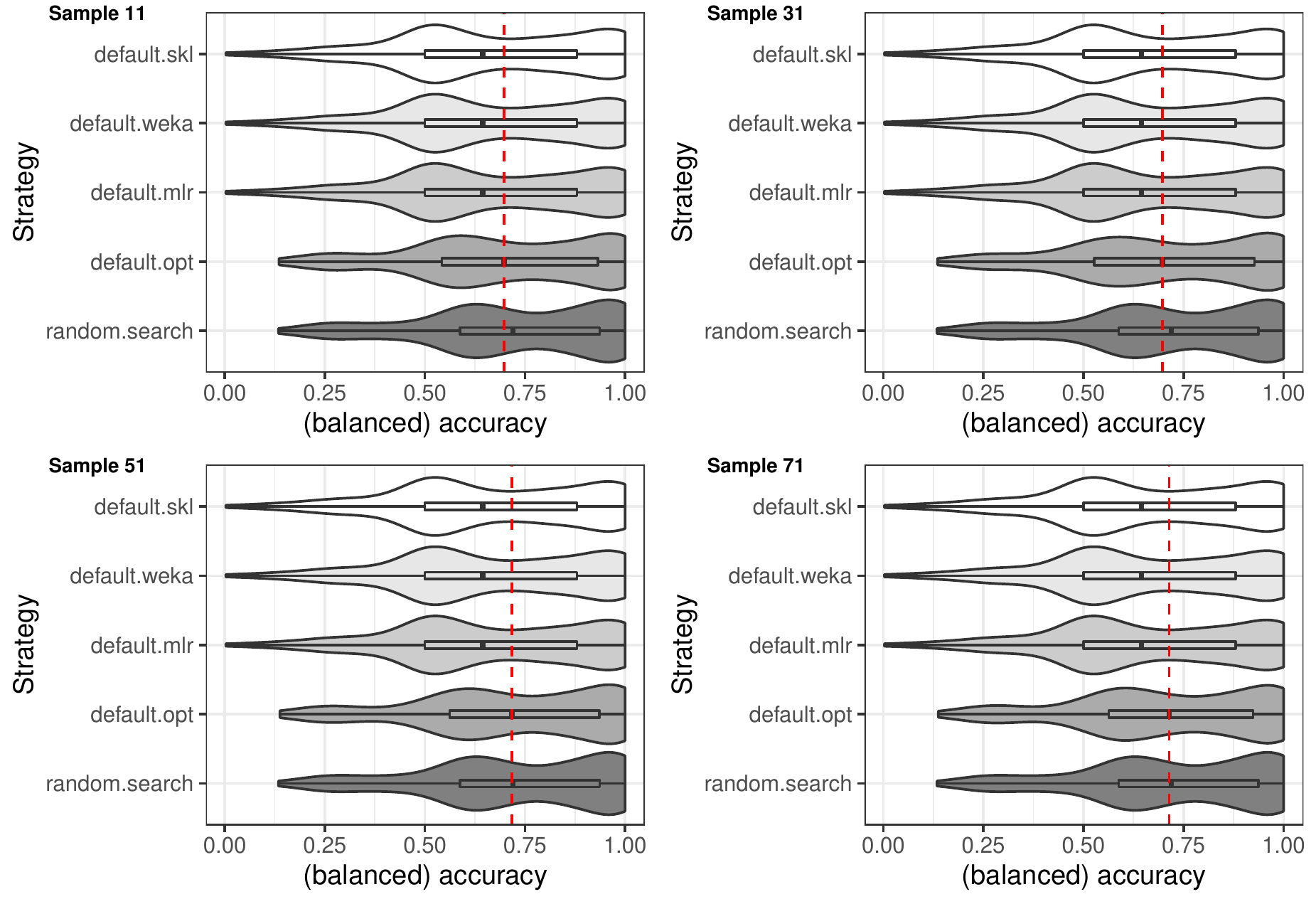}
    \caption{\acrshort{bac} performance distributions obtained by different \acrshort{hp} settings evaluated in test datasets with different sample sizes.}
    \label{fig:violin_performance_sample}
\end{figure}

{\normalsize 
The previous results lead us analyze this in more depth checking all the strategy distributions for each dataset sample size. Figure~\ref{fig:violin_performance_sample} now depicts four violin plots, each for a specific sample size: top charts show results for the optimization process considering $11$ and $31$ training datasets, while the bottom charts show results for $51$ and $71$ datasets.}


{\normalsize 
In general, the same behavior previously observed in the overall analysis is shown in all these graphs: \acrshort{rs} is the best-ranked strategy, followed by the new optimized settings, and the defaults from the used tools. However, looking at these results carefully bring us some new clues. 
When the sample size = $51$, the median of the best-optimized settings and the \acrshort{rs} median are very similar ($0.718$, $0.719$). This can be most easily seen when the red vertical line is used as a guide.
However, for a larger number of datasets (sample size = $71$), the performance of the default optimized settings drops again, increasing the difference between the \acrshort{rs} median ($0.713$ and $0.719$, respectively).
Hence, using as many training datasets as possible did not positively affect the results but increased the amount of time required to generate the settings.}


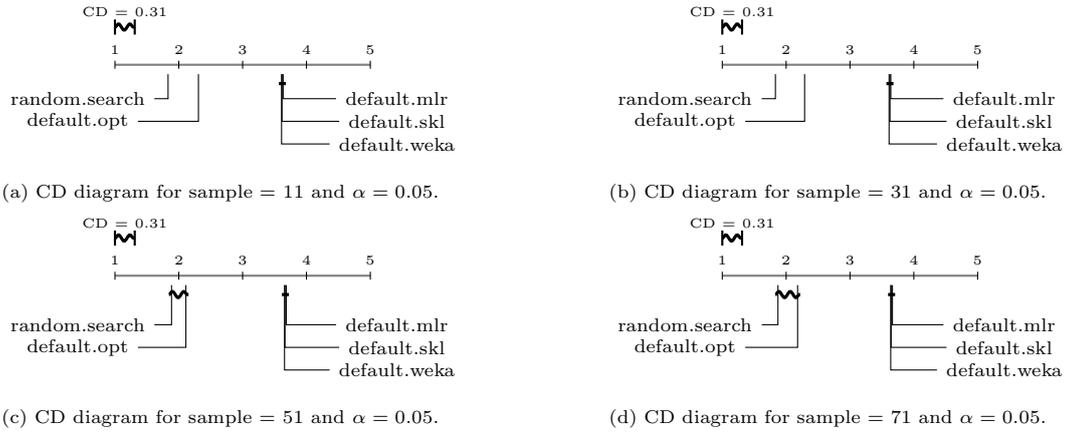
\begin{figure}[h!]
    \centering
    \centering
     \begin{subfigure}{0.35\textwidth}
            \begin{tikzpicture}[xscale=0.7]
                \node (Label) at (1.3860480558075452, 0.7){\tiny{CD = 0.31}}; 
                \draw[decorate,decoration={snake,amplitude=.4mm,segment length=1.5mm,post length=0mm},very thick, color = black] (1.2,0.5) -- (1.5720961116150902,0.5);
                \foreach \x in {1.2, 1.5720961116150902} \draw[thick,color = black] (\x, 0.4) -- (\x, 0.6);
                 
                \draw[gray, thick](1.2,0) -- (6.0,0); 
                \foreach \x in {1.2,2.4,3.6,4.8,6.0} \draw (\x cm,1.5pt) -- (\x cm, -1.5pt);
                \node (Label) at (1.2,0.2){\tiny{1}};
                \node (Label) at (2.4,0.2){\tiny{2}};
                \node (Label) at (3.6,0.2){\tiny{3}};
                \node (Label) at (4.8,0.2){\tiny{4}};
                \node (Label) at (6.0,0.2){\tiny{5}};
                \draw[decorate,decoration={snake,amplitude=.4mm,segment length=1.5mm,post length=0mm},very thick, color = black](4.281782945736434,-0.25) -- (4.4112403100775195,-0.25);
                \node (Point) at (2.1984496124031008, 0){};\node (Label) at (0.5,-0.45){\scriptsize{random.search}}; \draw (Point) |- (Label);
                \node (Point) at (2.7689922480620153, 0){};\node (Label) at (0.5,-0.75){\scriptsize{default.opt}}; \draw (Point) |- (Label);
                \node (Point) at (4.36124031007752, 0){};\node (Label) at (6.5,-0.45){\scriptsize{default.mlr}}; \draw (Point) |- (Label);
                \node (Point) at (4.339534883720931, 0){};\node (Label) at (6.5,-0.75){\scriptsize{default.skl}}; \draw (Point) |- (Label);
                \node (Point) at (4.331782945736434, 0){};\node (Label) at (6.5,-1.05){\scriptsize{default.weka}}; \draw (Point) |- (Label);
            \end{tikzpicture}
        
          \caption{CD diagram for sample = 11 and $\alpha=0.05$.}
         \label{fig:sample_11_cd_005}
    \end{subfigure}
     \hspace{2cm}
    \begin{subfigure}{0.35\textwidth}
        \begin{tikzpicture}[xscale=0.7]
            \node (Label) at (1.3860480558075452, 0.7){\tiny{CD = 0.31}}; 
                \draw[decorate,decoration={snake,amplitude=.4mm,segment length=1.5mm,post length=0mm},very thick, color = black] (1.2,0.5) -- (1.5720961116150902,0.5);
                \foreach \x in {1.2, 1.5720961116150902} \draw[thick,color = black] (\x, 0.4) -- (\x, 0.6);
                 
                \draw[gray, thick](1.2,0) -- (6.0,0); 
                \foreach \x in {1.2,2.4,3.6,4.8,6.0} \draw (\x cm,1.5pt) -- (\x cm, -1.5pt);
                \node (Label) at (1.2,0.2){\tiny{1}};
                \node (Label) at (2.4,0.2){\tiny{2}};
                \node (Label) at (3.6,0.2){\tiny{3}};
                \node (Label) at (4.8,0.2){\tiny{4}};
                \node (Label) at (6.0,0.2){\tiny{5}};
                \draw[decorate,decoration={snake,amplitude=.4mm,segment length=1.5mm,post length=0mm},very thick, color = black](4.2848837209302335,-0.25) -- (4.417441860465115,-0.25);
                \node (Point) at (2.1984496124031008, 0){};\node (Label) at (0.5,-0.45){\scriptsize{random.search}}; \draw (Point) |- (Label);
                \node (Point) at (2.7472868217054263, 0){};\node (Label) at (0.5,-0.75){\scriptsize{default.opt}}; \draw (Point) |- (Label);
                \node (Point) at (4.3674418604651155, 0){};\node (Label) at (6.5,-0.45){\scriptsize{default.mlr}}; \draw (Point) |- (Label);
                \node (Point) at (4.351937984496124, 0){};\node (Label) at (6.5,-0.75){\scriptsize{default.skl}}; \draw (Point) |- (Label);
                \node (Point) at (4.334883720930233, 0){};\node (Label) at (6.5,-1.05){\scriptsize{default.weka}}; \draw (Point) |- (Label);
        \end{tikzpicture}
        \caption{CD diagram for sample = 31 and $\alpha=0.05$.}
        \label{fig:sample_31_cd_005}
   
    \end{subfigure}
    \\
    \begin{subfigure}{0.35\textwidth}
        \begin{tikzpicture}[xscale=0.7]
            \node (Label) at (1.3860480558075452, 0.7){\tiny{CD = 0.31}}; 
            \draw[decorate,decoration={snake,amplitude=.4mm,segment length=1.5mm,post length=0mm},very thick, color = black] (1.2,0.5) -- (1.5720961116150902,0.5);
            \foreach \x in {1.2, 1.5720961116150902} \draw[thick,color = black] (\x, 0.4) -- (\x, 0.6);
             
            \draw[gray, thick](1.2,0) -- (6.0,0); 
            \foreach \x in {1.2,2.4,3.6,4.8,6.0} \draw (\x cm,1.5pt) -- (\x cm, -1.5pt);
            \node (Label) at (1.2,0.2){\tiny{1}};
            \node (Label) at (2.4,0.2){\tiny{2}};
            \node (Label) at (3.6,0.2){\tiny{3}};
            \node (Label) at (4.8,0.2){\tiny{4}};
            \node (Label) at (6.0,0.2){\tiny{5}};
            \draw[decorate,decoration={snake,amplitude=.4mm,segment length=1.5mm,post length=0mm},very thick, color = black](2.2135658914728684,-0.25) -- (2.5833333333333335,-0.25);
            \draw[decorate,decoration={snake,amplitude=.4mm,segment length=1.5mm,post length=0mm},very thick, color = black](4.337596899224806,-0.25) -- (4.47015503875969,-0.25);
            \node (Point) at (2.2635658914728682, 0){};\node (Label) at (0.5,-0.65){\scriptsize{random.search}}; \draw (Point) |- (Label);
            \node (Point) at (2.5333333333333337, 0){};\node (Label) at (0.5,-0.95){\scriptsize{default.opt}}; \draw (Point) |- (Label);
            \node (Point) at (4.42015503875969, 0){};\node (Label) at (6.5,-0.65){\scriptsize{default.mlr}}; \draw (Point) |- (Label);
            \node (Point) at (4.395348837209302, 0){};\node (Label) at (6.5,-0.95){\scriptsize{default.skl}}; \draw (Point) |- (Label);
            \node (Point) at (4.387596899224806, 0){};\node (Label) at (6.5,-1.25){\scriptsize{default.weka}}; \draw (Point) |- (Label);

        \end{tikzpicture}
         \caption{CD diagram for sample = 51 and $\alpha=0.05$.}
        \label{fig:sample_51_cd_005_2}
     
    \end{subfigure}
    \hspace{2cm}
    \begin{subfigure}{0.35\textwidth}    
    
        \begin{tikzpicture}[xscale=0.7]
            \node (Label) at (1.3860480558075452, 0.7){\tiny{CD = 0.31}}; 
            \draw[decorate,decoration={snake,amplitude=.4mm,segment length=1.5mm,post length=0mm},very thick, color = black] (1.2,0.5) -- (1.5720961116150902,0.5);
            \foreach \x in {1.2, 1.5720961116150902} \draw[thick,color = black] (\x, 0.4) -- (\x, 0.6);
             
            \draw[gray, thick](1.2,0) -- (6.0,0); 
            \foreach \x in {1.2,2.4,3.6,4.8,6.0} \draw (\x cm,1.5pt) -- (\x cm, -1.5pt);
            \node (Label) at (1.2,0.2){\tiny{1}};
            \node (Label) at (2.4,0.2){\tiny{2}};
            \node (Label) at (3.6,0.2){\tiny{3}};
            \node (Label) at (4.8,0.2){\tiny{4}};
            \node (Label) at (6.0,0.2){\tiny{5}};
            \draw[decorate,decoration={snake,amplitude=.4mm,segment length=1.5mm,post length=0mm},very thick, color = black](2.196511627906977,-0.25) -- (2.665503875968992,-0.25);
            \draw[decorate,decoration={snake,amplitude=.4mm,segment length=1.5mm,post length=0mm},very thick, color = black](4.315891472868217,-0.25) -- (4.448449612403101,-0.25);
            \node (Point) at (2.246511627906977, 0){};\node (Label) at (0.5,-0.65){\scriptsize{random.search}}; \draw (Point) |- (Label);
            \node (Point) at (2.6155038759689924, 0){};\node (Label) at (0.5,-0.95){\scriptsize{default.opt}}; \draw (Point) |- (Label);
            \node (Point) at (4.398449612403101, 0){};\node (Label) at (6.5,-0.65){\scriptsize{default.mlr}}; \draw (Point) |- (Label);
            \node (Point) at (4.373643410852713, 0){};\node (Label) at (6.5,-0.95){\scriptsize{default.skl}}; \draw (Point) |- (Label);
            \node (Point) at (4.365891472868217, 0){};\node (Label) at (6.5,-1.25){\scriptsize{default.weka}}; \draw (Point) |- (Label);
        \end{tikzpicture}
        \caption{CD diagram for sample = 71 and $\alpha=0.05$.}
         \label{fig:sample_71_cd_005}
    \end{subfigure}
 
    \caption[Comparison of the \acrshort{bac} values of the HP setting strategies for \acrshortpl{svm} according to the Nemenyi test with $\alpha = 0.05$.]{Comparison of the \acrshort{bac} values of the HP setting strategies for \acrshortpl{svm} according to the Nemenyi test with $\alpha = 0.05$. Groups of strategies that are not significantly different are connected.}
    
    \label{fig:cd_diagram_tree}
     
    \label{fig:nemenyi_samples}
  
\end{figure}

{\normalsize 
Figure~\ref{fig:nemenyi_samples} shows the Friedman-Nemenyi results, with $\alpha = 0.05$, when comparing strategies considering different datasets' sample sizes separately. The \acrshort{cd} diagram at the top show the results for sizes = \{11, 31\}, while the bottom ones show for sizes = \{51, 71\}. 
For all cases, 
\acrshort{rs} was the best-ranked whereas our strategy (default.opt) was the second one, both significantly
outperforming the traditional defaults.
In addition, it can be observed that our strategy does not present statistically significant differences to the tuning technique when the sample size = \{51, 71\}, i.e., using the new optimized settings led to equivalent results when compared to \acrshort{rs}.
}

{\normalsize 
Considering Figure~\ref{fig:nemenyi_samples}, especially the results for the 51 and 71 size samples, we confirm our hypothesis that a small set of new optimized \acrshort{hp} settings can improve predictive performance compared to traditional \acrshort{hp} default settings. Notably, our strategy is even comparable to a tuning technique, but with a much lower computational cost.
}


\end{document}